\documentclass[sigconf]{acmart}

\usepackage{pdfpages}
\usepackage{bm}
\usepackage{amsmath}
\usepackage{graphicx}
\usepackage{multirow}
\usepackage[justification=centering]{caption}

\AtBeginDocument{%
  \providecommand\BibTeX{{%
    \normalfont B\kern-0.5em{\scshape i\kern-0.25em b}\kern-0.8em\TeX}}}

\copyrightyear{2022}
\acmYear{2022}
\setcopyright{acmcopyright}
\acmConference[CIKM '22] {Proceedings of the 31st ACM International Conference on Information and Knowledge Management}{October 17--21, 2022}{Atlanta, GA, USA.}
\acmBooktitle{Proceedings of the 31st ACM International Conference on Information and Knowledge Management (CIKM '22), October 17--21, 2022, Atlanta, GA, USA}
\acmPrice{15.00}
\acmISBN{978-1-4503-9236-5/22/10}
\acmDOI{10.1145/3511808.3557359}

\usepackage{amssymb}
\usepackage{balance}
\usepackage{booktabs}
\usepackage{graphicx}
\usepackage{pifont}
\newcommand{\cmark}{\ding{51}}%
\newcommand{\xmark}{\ding{55}}%
\DeclareMathAlphabet{\mathcal}{OMS}{cmsy}{m}{n}
\newcommand*{\tabref}[1]{\tablename~\ref{#1}}
\newcommand*{\figref}[1]{\figurename~\ref{#1}}

\begin{document}

\title{Improving Personality Consistency in Conversation by Persona Extending}


\author{Yifan Liu}
\affiliation{%
  \institution{CCIIP Laboratory, School of Computer Science and Technology, Huazhong University of Science and Technology}
  \institution{Joint Laboratory of HUST and Pingan Property \& Casualty Research (HPL)}
  \country{China}
}
\email{yifaan@hust.edu.cn}

\author{Wei Wei}
\authornote{Corresponding author.}
\affiliation{%
  \institution{CCIIP Laboratory, School of Computer Science and Technology, Huazhong University of Science and Technology}
  \institution{Joint Laboratory of HUST and Pingan Property \& Casualty Research (HPL)}
  \country{China}
}
\email{weiw@hust.edu.cn}

\author{Jiayi Liu}
\affiliation{%
  \institution{Alibaba group}
  \country{China}
}
\email{ljy269999@alibaba-inc.com}

\author{Xianling Mao}
\affiliation{%
  \institution{Beijing Institute of Technology}
  \country{China}
}
\email{maoxl@bit.edu.cn}

\author{Rui Fang}
\affiliation{%
  \institution{Ping An Property \& Casualty Insurance company of China, Ltd}
  \country{China}
}
\email{fangrui051@pingan.com.cn}

\author{Dangyang Chen}
\affiliation{%
  \institution{Ping An Property \& Casualty Insurance company of China, Ltd}
  \country{China}
}
\email{chendangyang273@pingan.com.cn}

\renewcommand{\shortauthors}{Liu and Wei, et al.}

\begin{abstract}
  Endowing chatbots with a consistent personality plays a vital role for agents to deliver human-like interactions. However, existing personalized approaches commonly generate responses in light of static predefined personas depicted with textual description, 
  which may severely restrict the interactivity of human and the chatbot, especially when the agent needs to answer the query excluded in the predefined personas, which is so-called out-of-predefined persona problem (named \textbf{OOP} for simplicity).
  To alleviate the problem, in this paper we propose a novel \emph{retrieval-to-prediction} paradigm consisting of two subcomponents, 
  namely, (1) \textbf{\emph{\underline{P}ersona \underline{R}etrieval \underline{M}odel (PRM)}}, it retrieves a persona from a global collection based on a Natural Language Inference (NLI) model, the inferred persona is consistent with the predefined personas; 
  and (2) \textbf{\emph{\underline{P}osterior-\underline{s}cored \underline{Transformer} (PS-Transformer)}}, it adopts a persona posterior distribution that further considers the actual personas used in the ground response, maximally mitigating the gap between training and inferring. 
  Furthermore, we present a dataset called IT-ConvAI2 that first highlights the OOP problem in personalized dialogue. Extensive experiments on both IT-ConvAI2 and ConvAI2 demonstrate that our proposed model yields considerable improvements in both automatic metrics and human evaluations. 
  All the data and codes are publicly available at \url{https://github.com/CCIIPLab/Persona_Extend/}.
\end{abstract}



\maketitle

\section{Introduction}


\begin{figure}[tb]
  \centering
  \includegraphics{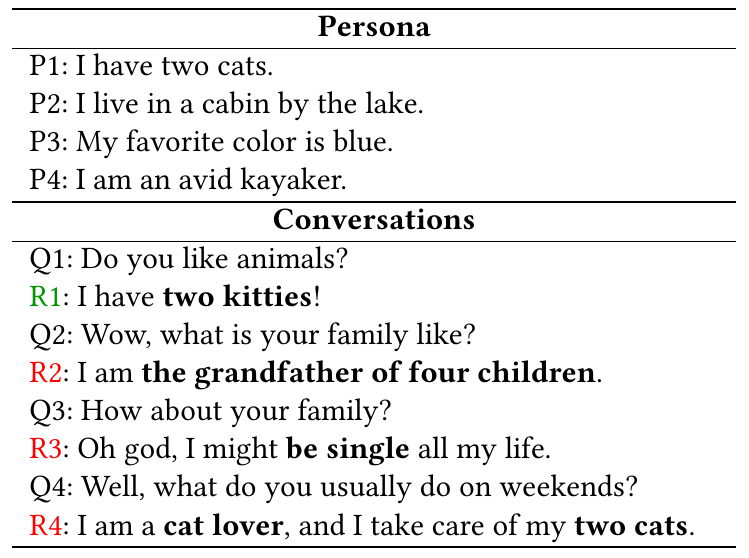}
  \captionsetup{justification=justified}
  \caption{Examples generated by \emph{Transformer} to illustrate following problems: (1) Response Conflict: \textcolor{red}{R2} \& \textcolor{red}{R3} fabricate different personas, which are conflict with each other. (2) Inappropriate Persona: P4 is more appropriate for answering Q4 than P1.}
  \label{table:example}
\end{figure}

Developing a more human-like dialogue system has been an important topic in artificial intelligence, where one of the major challenges is to maintain a consistent persona~\cite{li2016persona,qian2017assigning,mazare2018training,song2019exploiting,wolf2019transfertransfo,song2020generating,liu2020you,song2021bob}. Key-value lists are first used to construct structured profile explicitly, including name, gender, age, location, etc~\cite{qian2017assigning,zheng2019personalized,zheng2020pre}.
More recently, \citet{zhang2018personalizing} define the personality as several textual persona sentences as stated in \figref{table:example}. As the unstructured personas are natural, vivid, and facilitate the description of complicated personalities, it sparks a wide range of interest in developing generators of personality-consistent responses~\cite{fillwock2018identification}.
To enhance the understanding of predefined textual personas: 
\citet{wolf2019transfertransfo} first employ pretrained model that leveraged the general dialogue corpus to understand textual personas better.
\citet{song2021bob} pretrain an encoder with non-dialogue inference data to strengthen consistency understanding. 
\citet{xu2020neural} propose enriching predefined persona by searching related topics, and \citet{majumder2020like} generalize predefined personas by leveraging commonsense to guess the underlying personas.

However, there are several limitations for existing methods on generating responses based on textual persona sentences. 
\textbf{First}, 
most current methods adopt only the predefined personas for response genreation, and thus easily fail in generating the reasonable response if confronting the OOP problem.
As shown in \figref{table:example}, Q2 and Q3 are two classical OOP examples, which cannot directly answer the query like ``farmily sistutation'' with the given personas. 
However, without more external knowledge, 
the agent may fabricate several inappropriate personas (e.g., ``have four children'') that may be inconsistent with the prior persons, which is so-called personality inconsistent generation problem.
\textbf{Second}, 
although there exists several works on expanding predefined personas for generation, they merely focus on paraphrasing  a specific predefined,
without considering the consistency of the expanded persona with the given query and other predefined personas~\cite{xu2020neural, majumder2020like}. It may extend a persona that is not suitable for response, even inconsistent with the rest of personas, and lead to contradiction problems.
\textbf{Third}, 
some methods~\cite{zheng2019personalized,zheng2020pre} simply fuse all personas into the generation process cursorily, 
which may lead to the output of an inappropriate response with an inconsistent persona.
As shown in \figref{table:example}, 
such as ``I am kayaker'' may be more relevant to Q4, however the agent still graft ``I have two cats'', as ``I have pets'' is a more general persona in the whole dataset, as compared to ``I am kayaker'', which is so-called long-tail bias problem. 
Under such circumstance, it is non-trivial  to directly solve the OOP problem.

In this paper, we argue the importance of addressing the OOP problem, which may significantly improve the consistency of existing personalized dialogue systems. 
Recall the examples shown in \figref{table:example}, for the OOP queries (e.g., Q2 and Q3), an reasonable solution is to obtain an appropriate persona from an external knowledge based on the per-defined personas.
However, the generator may overlook the appropriate persona we expand (e.g., R4), so we must filter the existing textual personas before generating responses. 
Therefore, we design a pipeline that retrieves persona and selects persona for addressing the OOP problem.
Inspired by this, our research starts by asking: \textbf{What is the principle of retrieving persona for OOP query?}
Here, the first important issue is whether the retrieved personas are semantically consistent with the predefined ones.
For example, ``I don't like pets'' obviously implies ``I don't have a dog or a cat''.
Therefore, the retrieved personas should be compared with the predefined ones for semantic conflict checking. 
The second question is: \textbf{How to ensure that endowing the chatbot (e.g., the generated response) with the retrieved persona?}
Generally, the existing generative models trend to select commonly appeared personas for generation. With the target of avoiding the general response generation, the generation model cannot use all of retrieved personas. 
Instead, we encourage the model to select the most query-relevant persona before generation, significantly improving the relevance of the generated response to the context.

Therefore, this paper proposes a novel retrieval-to-prediction pipeline consisting of \emph{PRM} and \emph{PS-Transformer}. Specifically, \emph{PRM} is designed as a ranking module that extends personas by retrieving from a global persona set\footnote{In this paper we simply take all personas from the test set of ConvAI2 as our global personas.}.
In particular, we leverage Natural Language Inference (NLI) to select personas that do not conflict with predefined personas. 
\emph{PS-Transformer} adopts \emph{Target-Guided Persona Scorer} to predict the availabilities of each persona to the query by posterior information. Incorporated with such a persona distribution, our proposed model is able to select the most suitable persona to generate responses.
We build a challenging set named \underline{I}nadequate-\underline{T}iny-\underline{ConvAI2} (IT-ConvAI2) by removing those query-related personas from the original ConvAI2 dataset.
In this way, we verify that the \emph{PRM} could steadily extend a suitable new persona to tackle the OOP problem
and facilitate \emph{PS-Transformer} to generate personality-consistent responses.
On both IT-ConvAI2 and ConvAI2, 
we demonstrate that our method directly improves the coherence of generation at the personality level. 

The main contributions of this research are summarized:

\textbf{First}, we propose a novel framework solving the OOP problem in dialogue generation. This framework involves two processes, i.e., conflict-detecting persona retrieving and dialogue generation with selected personas.

\textbf{Second}, we are the first to leverage NLI to estimate the coherence from persona candidates to predefined personas. Extensive experiments demonstrate that our proposed \emph{PRM} can gather better personas than others.

\textbf{Third}, we propose a novel \emph{PS-Transformer} introducing the \emph{Target-Guided Persona Scorer} to predict persona distributions instead of fusing them roughly. The \emph{PS-Transformer} yields the best results on both IT-ConvAI2 and ConvAI2.

\section{Related Work}

\subsection{Personalized Dialogue}

Although neural response generation models have achieved promising results \cite{vinyals2015neural, shang2015neural, hochreiter1997long, zhang2020dialogpt, zhao2017learning, target2021wei, emotion2019wei}, they are still unsatisfactory. 
Previous work~\cite{eggins2005analysing} investigated that topic changing will significantly satisfy conversational participants. Furthermore, \citet{mitsuda2019information} proposed that $78.5$\% of the perceived information during chit-chat is directly related to personal information. \citet{li2016persona} first proposed a personalized dialogue system to introduce personal information into dialogue generation. 
After this, \citet{qian2017assigning} proposed WD Profile Dataset, and \citet{zhang2018personalizing} proposed ConvAI2. Such personalized dialogue contributed to the development of both retrieval-based and generative-based personalized dialogue models.

In the line of retrieval-based methods \cite{zhang2018personalizing,gu2019dually,gu2021partner}, \citet{gu2021partner} found it is helpful to utilize personas in response selection. 
Although our proposed \emph{PRM} retrieves personas, our pipeline method does not belong to the retrieval-based methods. Because our method does not directly take the retrieval results as responses, but uses them as the basis for generating, which facilitates the generation of informative and consistent responses.

In the line of generative-based methods, 
\citet{li2016persona} first took user embedding as an implicit persona in multi-turn dialogues. However, it relied on expensive speak-tagged dialogue data.
Recent works incorporated explicit persona into the generation in two ways:
(1) \citet{qian2017assigning} and \citet{zheng2020pre} defined personality as structured key-value profiles consisting of some basic personal information such as name, age, and location. 
(2) \citet{zhang2018personalizing} contributed a chat-oriented dataset, taking personality as a predefined collection of textually described persona sentences. 
Most of the persona dialogue methods\cite{zhang2018personalizing, wolf2019transfertransfo, song2019exploiting, song2020generating, yavuz2019deepcopy, song2021bob} focused on how to understand personas better in the latter high-quality corpus.
Specifically, 
\citet{zhang2018personalizing} employed basic Seq2Seq splicing personas with the query without distinguishing them. 
\citet{wolf2019transfertransfo} first introduced transfer learning by fine-tuning pretrained model to improve the quality of generation.
However, all methods above take the agent's personality as a predefined closed set. Once the query goes beyond predefined personas (OOP problem), the agent tends to fabricate a new persona, resulting in a risk of inconsistent personality. To tackle the problem, we propose our retrieval-to-prediction pipeline that extends persona before generation.

\subsection{Natural Language Inference}
The task of Natural Language Inference (NLI) is to learn a function $f_{\mathrm{NLI}}(p,h)=\{\mathrm{E}, \mathrm{N}, \mathrm{C}\}$, where $p$ and $h$ denote premise and hypothesis respectively. The outputs E, N and C represent the conjunction, neural and contradiction relations between premises and hypotheses.
Since the release of the large-scale corpus SNLI \cite{bowman2015large}, deep neural network approaches have made promising progress\cite{chen2017enhanced,gong2018natural,kim2019semantic}.
\citet{welleck2019dialogue} modeled the detection of conversational consistency as an NLI task and proposed the Dialogue NLI dataset.
And \citet{song2020generating} adopted the RL framework to leverage NLI knowledge as a reward. 
\citet{song2021bob} further pretrained on NLI task to ensure generating responses that entail predefined personas.

Motivated by this, we argue that NLI is crucial for personal retrieval to identify the relevances between persona candidates and predefined personas.
So we consider the entail and conflict with the predefined personas in the NLI perspective when \emph{PRM} retrieves the persona, thus providing suitable persona for the generative model.

\subsection{Knowledge Enhanced Dialogue}
The incorporation of knowledge has been shown to be an effective way to improve the performance of dialogue generation. There is a trend to leverage many domain-specific knowledge bases to ground neural models \cite{xu2017incorporating,zhu2017flexible,gu2016incorporating,zou2021multi,zhao22multiview,pan2021context}, in which the textual persona sentences are one of the most frequently considered knowledge \cite{lian2019learning}.
Recently, \citet{lian2019learning} propose that compared with the knowledge posterior distribution that further considers the actual knowledge used in real responses, the prior distribution has a large variance, and therefore, it is difficult for existing models to simply select the appropriate knowledge based on the prior distribution during training. On this basis, \citet{song2019exploiting} and \citet{gu2019dialogwae} use posterior distributions effectively to ensures that knowledge is better utilized in generating responses.

We borrow the idea that leverages posterior distribution to select the appropriate knowledge with several differences in motivation and methodology: (1) We use the posterior distribution to select the actual personas rather than traditional knowledge in the grounded response. (2) Compared to fusing all personas into one representation~\cite{gu2019dialogwae}, we consider the modeling of persona selection distribution.

\begin{figure}[tb]
  \centering
  \includegraphics[width=150px]{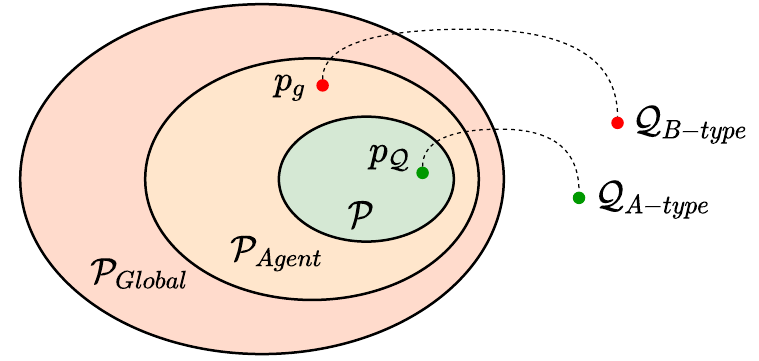}
  \caption{Relations between personas and queries.}
  \label{fig:case}
\end{figure}

\begin{figure}[tb]
  \centering
  \includegraphics[width=240px]{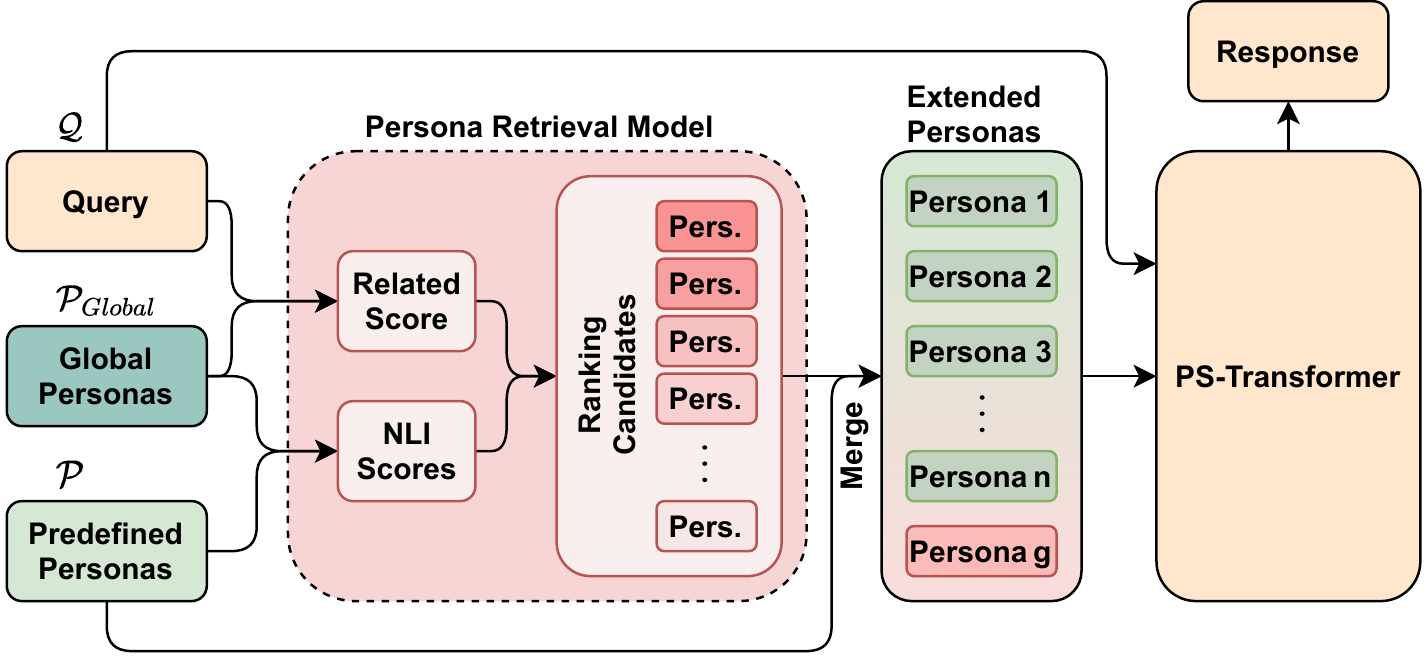}
  \caption{Overview of the retrieval-to-prediction paradigm.}
  \label{fig:overview}
\end{figure}

\section{Methodology}\label{sec:method}

\subsection{Task Definition}\label{sec:taskdef}

In this paper, our personalized dialogue generation problem aims at endowing a dialogue system with a consistent personality  for building a human-like conversation sysmte, which can be formally defined as follows,  
given a query $\mathcal{Q} = \{q_i\}_{i=1}^{m}$ and a set of predefined personas $\mathcal{P} = \{p_1, p_2, \dots, p_{n}\}$, where each persona depicted with a sentence $p_i = \{w_j\}_{j=1}^{m} (i \in \{1, 2, \dots, n\})$, the task aims to generate a response $\mathcal{R} = \{r_i\}_{i=1}^m$ coherent to both the query and agent's personas.

As stated in \figref{fig:case}, assuming a global persona collection $\mathcal{P}_{Global}$ for all agents, personas belonging to a specific agent could be declared as $\mathcal{P}_{Agent}\ (\mathcal{P}_{Agent} \subsetneqq \mathcal{P}_{Global})$. Usually, we handly predefine a persona set $\mathcal{P}\ (\mathcal{P} \subsetneqq \mathcal{P}_{Agent})$ for the agent.
On this assumption, we divide queries into A-type and B-type. 
A-type queries can be answered based on predefined $p_{\mathcal{Q}}\ (p_\mathcal{Q} \in \mathcal{P})$, while B-type queries need us to detect a new persona $p_g\ (p_g \in \mathcal{P}_{Agent},\ p_g \notin \mathcal{P})$ to tackle.

To simplify the problem, we assume that global persona set $P_{Global}$ must contain at least one persona related to the query. So this paper focuses on retrieving a suitable persona $p_g$ from $P_{Global}$ and generating responses coherent with extended personas. The personalized dialogue generation can be briefly stated below:
\begin{equation}
\begin{aligned}
& \mathbf{P} (\mathcal{R}|\mathcal{Q}, \mathcal{P}, \mathcal{P}_{Global}) \\
= & \mathbf{P} (\mathcal{R}|\mathcal{Q}, \mathcal{P} \cup \{p_g\}) \cdot \mathbf{P}(p_g | \mathcal{Q}, \mathcal{P}, \mathcal{P}_{Global}), \\
\end{aligned}
\end{equation}
where $\mathbf{P}(p_g | \mathcal{Q}, \mathcal{P}, \mathcal{P}_{Global})$ denotes detecting a new persona $p_g$ from $\mathcal{P}_{Global}$ considering current query $\mathcal{Q}$ and predefined personas $\mathcal{P}$.
And $\mathbf{P}(\mathcal{R}|\mathcal{Q}, \mathcal{P} \cup \{p_g\}) = \prod_{t=1}^{n_r} \mathbf{P}(r_t|\mathcal{Q}, \mathcal{P} \cup \{p_g\}, r_{<t})$ represents the response generation based on both the context query $\mathcal{Q}$ and extended personas $\mathcal{P}\cup \{p_g\}$.

\begin{figure*}[htbp]
  \centering
  \includegraphics[width=450px]{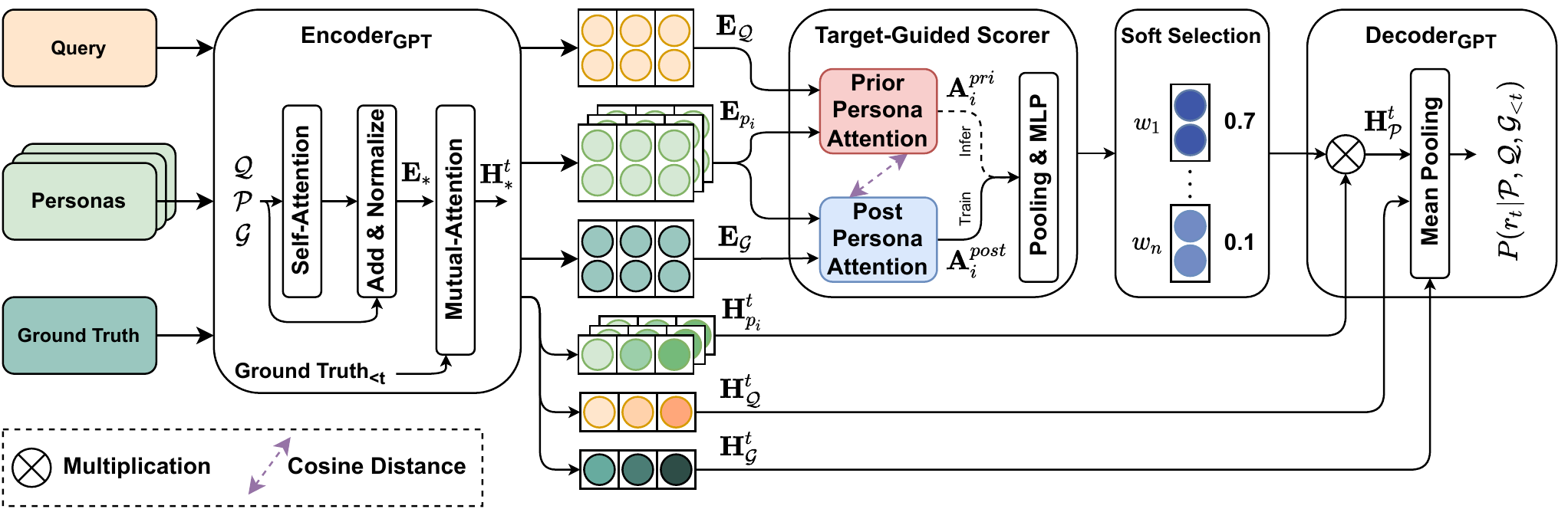}
  \captionsetup{justification=justified}
  \caption{The neural network architecture of the proposed \emph{PS-Transformer}.}
  \label{fig:generator}
\end{figure*}

\subsection{Overview}\label{sec:overview}

As stated in \figref{fig:overview}, we designed a retrieval-to-prediction pipeline that combines persona extending and response generation. The pipeline consists of two stages: 
\emph{PRM} retrieves a persona from the global persona set based on the predefined personas and the context query.
Then \emph{PS-Transformer} generates a response with the query and the extended personas. The details of our method will be explained in Section~\ref{sec:retriever} and Section~\ref{sec:generator}, respectively.

\subsection{Persona Retrieval Model}\label{sec:retriever}

The \emph{Persona Retrieval Model (PRM)} is responsible for addressing $\bm{P}(p_g | \mathcal{Q}, \mathcal{P}, \mathcal{P}_{Global})$, i.e. ranking all the candidate personas and picking the mostly one for the agent. 
Firstly we handly prepare a collection of persona candidates $\mathcal{P}_{Global}$\footnote{
  To avoid the label leaking, 
  we make sure that all candidates did not be used for training our sentence-pair model and NLI model.}
from the ConvAI2 dataset. As stated in \figref{fig:overview}, the \emph{PRM} ranks all the candidates based on both its relanvences to query and predefined personas.

We employ a sentence-pair matching model to estimate the logical association between query and persona candidate. The score predicted by the binary classification model is the query-persona relevance, as stated in Equation \ref{eq:relevance}.
\begin{equation}\label{eq:relevance}
\begin{aligned}
r & = \mathrm{Related}(\mathcal{Q}, p_g)
\end{aligned}
\end{equation}

\citet{song2020generating} finds that NLI models could be used to calculate the coherence between response and query. Inspired by their work, We empirically adopt a standard pretrained NLI model \cite{gao2021adapting} to check textual entailment and conflict between the persona candidate and predefined personas.
We apply the maximum algorithm to encourage the persona candidate $p_g$ closing to predefined personas $\mathcal{P}$, as shown in Equation~\ref{eq:entail}.
\begin{equation}\label{eq:entail}
\begin{aligned}
e & = \mathrm{Entail}(\mathcal{P}, p_g) \\
& = \max \limits_{p_i \in \mathcal{P}}\{\mathrm{Entail}(p_i, p_g)\}
\end{aligned}
\end{equation}
where $\mathrm{Entail}(\cdot,\cdot)$ is the entailment score predicted by our model. 
The persona candidate $p_g$ should be punished if it conflicts with any predefined persona of $\mathcal{P}$, so we also apply the maximum to calculate the conflict score for persona candidate $p_g$ in Equation~\ref{eq:conflict}.
\begin{equation}\label{eq:conflict}
\begin{aligned}
c & = \mathrm{Conflict}(\mathcal{P}, p_g) \\
& = \max \limits_{p_i \in \mathcal{P}}\{\mathrm{Conflict}(p_i, p_g)\}
\end{aligned}
\end{equation}
where $\mathrm{Conflict}(\cdot,\cdot)$ is the conflict score given by our model. 

We propose two approaches to combine scores $r,e,c$ as our ranking methods:

\begin{enumerate}

\item \textbf{Heuristic Rules} (NLI\textsubscript{HR}): We first retrieve top-10 candidates from $\mathcal{P}_{Global}$ with the highest $r$ score, and these persona candidates should be most relevant to the query $\mathcal{Q}$. Then, we take the persona with the highest $e$ score from the top-3 lowest $c$ scores to
encourage both its low conflict and strong entailment to predefined personas.

\item \textbf{Weight Combination} (NLI\textsubscript{WC}): We adopt three regulator $\alpha$, $\beta$, and $\gamma$ to construct a combined score $S=\alpha\cdot r + \beta\cdot (1-c) + \gamma\cdot e$. Then we sort the candidates with $S$ scores in descending order and take the first one as a result. In this paper we set $\alpha =0.75, \beta=0.25, \gamma=0.10$.

\end{enumerate}

\subsection{Posterior-scored Transformer}\label{sec:generator}

The dialogue generator we proposed is a transformer-based model stated in \figref{fig:generator}. Following the champion model in the ConvAI2 competition~\cite{dinan2020second}, we adopt OpenAI GPT~\cite{radford2018improving} as our weight-shared encoder $\mathrm{Encoder_{GPT}}(\cdot)$ and decoder $\mathrm{Decoder_{GPT}}(\cdot)$. 

\subsubsection{\textbf{Target-Guided Persona Scorer}}\label{sec:tgp}

Let $\mathcal{Q}, p_i, \mathcal{G}$ denote query, $\mathrm{i^{th}}$-persona, and ground truth (also known as target response), respectively. As stated in Equation~\ref{eq:encode}, we first adopt $\mathrm{Encoder_{GPT}(\cdot)}$ to turn the token-level embeddings into fixed-length representations at timestamp $t$.
\begin{equation}\label{eq:encode}
\begin{aligned}
\mathbf{E}_{\mathcal{Q}}, \mathbf{H}_{\mathcal{Q}}^{t} &= \mathrm{Encoder_{GPT}}(\mathcal{Q}) \\
\mathbf{E}_{\mathcal{G}}, \mathbf{H}_{\mathcal{G}}^{t} &= \mathrm{Encoder_{GPT}}(\mathcal{G}) \\
\mathbf{E}_{p_i}, \mathbf{H}_{p_i}^{t} &= \mathrm{Encoder_{GPT}}(p_i), i = 1,2,...,n_p 
\end{aligned}
\end{equation}
where $\mathbf{E}_*$ represents time-independent sentence embeddings of each input after self-attention only. $\mathbf{H}_*^t$ denotes the hidden states of each input after interacting with generated $\mathcal{G}_{<t}$.

The multi-head self-attention (denoted as $\mathrm{MHA}$, \cite{vaswani2017attention}) is used to compute
the importance from $\mathrm{i^{th}}$-persona to either query $\mathcal{Q}$ or the ground truth $\mathcal{G}$.
For each persona $p_i$ we calculate the attention $\mathbf{A}_i^*$ in Equation~\ref{eq:attn}.
\begin{equation}\label{eq:attn}
\begin{aligned}
\mathbf{A}_i^{pri} &= \mathrm{MHA}_{pri}(\mathbf{Q}=\mathbf{E}_{p_i}, \mathbf{K}=\mathbf{E}_{\mathcal{Q}}, \mathbf{V}=\mathbf{E}_{\mathcal{Q}})\\
\mathbf{A}_i^{post} &= \mathrm{MHA}_{post}(\mathbf{Q}=\mathbf{E}_{p_i}, \mathbf{K}=\mathbf{E}_{\mathcal{G}}, \mathbf{V}=\mathbf{E}_{\mathcal{G}})
\end{aligned}
\end{equation}

Since attention $\mathbf{A}_i^*$ denotes the importance of each persona to the response. A two-layer multilayer feedforward perceptron (MLP) with a sigmoid activation is used to turn them into a comprehensed weight as stated in Equation~\ref{eq:mlp}.
\begin{equation}\label{eq:mlp}
\begin{aligned}
w_i^{*} = \sigma (\mathrm{MLP}(\mathbf{A}_i^*)),\ (* = post\ \mathrm{or}\ pri)
\end{aligned}
\end{equation}

The binary cross entropy (BCE) loss is adopted to optimize the capture of weight $w_i^{post}$:
\begin{equation}\label{loss:bce}
\begin{aligned}
\mathcal{L}_1 = -[w_i \log w_i^{post} + (1-w_i) \log (1-w_i^{post})]
\end{aligned}
\end{equation}

Besides, the cosine embedding loss is used to gain both attentions from prior and posterior network as stated in Equation~\ref{loss:cos}.
\begin{equation}\label{loss:cos}
\begin{aligned}
\mathcal{L}_2=1-\mathrm{cos}(\mathbf{A}_i^{post},\mathbf{A}_i^{pri})
\end{aligned}
\end{equation}

\subsubsection{\textbf{Decoder for Weighted-sum Attentions}}\label{sec:decode}

Firstly, the representation $\mathbf{H}_{\mathcal{P}}$ for the predefined persona set $\mathcal{P}$ could be incorporated from $\mathbf{H}_{p_i}$ in Equation~\ref{eq:encode} based on $w_i^{post}$ given by Equation~\ref{eq:mlp}.

\begin{equation}\label{eq:aggp}
\begin{aligned}
\mathbf{H}_{\mathcal{P}}^{t} = \sum_{i=1}^{n_p} w_i^{post} \cdot \mathbf{H}_{p_i}^{t}
\end{aligned}
\end{equation}

To give consideration to both query and the past generated words,
In each timestamp $t$ of decoding, representations of query, personas and past generated words are treated equally. The prediction of word $r_t$ is stated in Equation~\ref{eq:decode}.
\begin{equation}\label{eq:decode}
\begin{aligned}
\mathbf{H}_{dec}^t &= \mathrm{mean}(\mathbf{H}_\mathcal{Q}^{t}, \mathbf{H}_\mathcal{P}^{t}, \mathbf{H}_{\mathcal{G}}^{t}) \\
r_t &= \mathrm{Decoder_{GPT}}(\mathbf{H}_{dec}^t)
\end{aligned}
\end{equation}
where $\mathrm{mean(\cdot)}$ denotes averaging given matrices by element.

In essence, the \emph{PS-Transformer} read the persona set $\mathcal{P}$ and the query ${\mathcal{Q}}$ to predict the target response $\mathcal{G}$. So we apply the negative log-lokeihood loss during training.
\begin{equation}
\begin{aligned}
\mathcal{L}_3 &=-\log \left(\bm{P}(\mathcal{G}|\mathcal{P}, \mathcal{Q})\right) \\
&=-\sum_{t=1}^{|\mathcal{G}|} \log (\bm{P}(r_{t}|\mathcal{P}, \mathcal{Q}, \mathcal{G}_{<t}))
\end{aligned}
\end{equation}

\subsubsection{\textbf{Inferrence Network}}\label{sec:infer}
Similar to Equation~\ref{eq:aggp}, the predefined personas are soft-selected by weighted summation based on the $w_i^{pri}$ predicted in Equation~\ref{eq:mlp}:
\begin{equation}
\begin{aligned}
\mathbf{H}_{\mathcal{P}}^{t} = \sum_{i=1}^{n_p} w_i^{pri} \cdot \mathbf{H}_{p_i}^{t}
\end{aligned}
\end{equation}

During decoding, the response is generated in a self-recursion way as stated in Equation~\ref{eq:infer}.
\begin{equation}\label{eq:infer}
\begin{aligned}
\mathbf{H}_{dec}^t &= \mathrm{mean}(\mathbf{H}_\mathcal{Q}^{t}, \mathbf{H}_\mathcal{P}^{t}, \mathbf{H}_{\mathcal{R}}^{t}) \\
r_t &= \mathrm{Decoder_{GPT}}(\mathbf{H}_{dec}^t)
\end{aligned}
\end{equation}

%
%
%
%
%
%
%
%
%
%
%
%
%
%

\begin{table*}[!th]
  \centering
  \caption{Automatic evaluation on IT-ConvAI2. In this evaluation, we adopt NLI\textsubscript{WC} in Section~\ref{sec:retriever} as the \emph{PRM}.}
    \begin{tabular}{cccccccccccccc}
    \toprule
    \multirow{3}[6]{*}{\textbf{Model}} & \multirow{3}[6]{*}{\textbf{Pretrained}} &       & \multicolumn{5}{c}{IT-ConvAI2}        &       & \multicolumn{5}{c}{IT-ConvAI2 with \emph{PRM}} \\
\cmidrule{4-8}\cmidrule{10-14}          &       &       & \textbf{Consist} &       & \multicolumn{3}{c}{\textbf{Quality}} &       & \textbf{Consist} &       & \multicolumn{3}{c}{\textbf{Quality}} \\
\cmidrule{4-4}\cmidrule{6-8}\cmidrule{10-10}\cmidrule{12-14}          &       &       & Entail &       & BLEU  & ROUGE & CIDEr &       & Entail &       & BLEU  & ROUGE & CIDEr \\
    \midrule
    Seq2Seq & \xmark     &       & 0.115 &       & 5.62  & 1.71  & 8.77  &       & 0.178 &       & 5.69  & 1.71  & 9.06 \\
    PerCVAE & \xmark     &       & 0.306 &       & 2.26  & 0.93  & 4.46  &       & 0.380 &       & 2.27  & 0.96  & 4.22 \\
    DialogWAE & \xmark     &       & 0.077 &       & 4.13  & 1.12  & 5.81  &       & 0.103 &       & 3.84  & 1.09  & 5.27 \\
    \midrule
    Transformer & \cmark     &       & 0.539 &       & 6.21  & 1.55  & 10.56 &       & 0.495 &       & 6.17  & 1.52  & 11.11 \\
    TransferTransfo & \cmark     &       & 0.546 &       & 5.12  & 1.34  & 13.23 &       & 0.645 &       & 5.18  & 1.36  & 12.85 \\
    BoB   & \cmark     &       & 0.505 &       & 5.39  & 1.43  & 11.39 &       & 0.628 &       & 5.35  & 1.40  & 10.74 \\
    \midrule
    \textbf{PS-Transformer } & \cmark     &       & \textbf{0.560} &       & \textbf{7.12} & \textbf{1.71} & \textbf{14.43} &       & \textbf{0.670} &       & \textbf{7.35} & \textbf{1.73} & \textbf{15.88} \\
    \bottomrule
    \end{tabular}%
  \label{tab:result_inde}%
\end{table*}%
\section{Experimental Setup}

\subsection{Research Questions} \label{sec:rq}
To fully demonstrate the superiority of our method, we conduct experiments to verify the following six research questions (RQ): 
\begin{itemize}
    \item (\textbf{RQ1}): Can our proposed pipeline, consisting of \emph{PRM} and \emph{PS-Transformer}, yield good results on automatic metrics in response to OOP queries? (See Section~\ref{sec:rq1})
    \item (\textbf{RQ2}): Can our proposed \emph{PRM} actually solve the OOP problem to some extent? Will the quality of the response generated by \emph{PS-Transformer} be better if we solved the OOP problem better? (See Section~\ref{sec:rq2})
    \item (\textbf{RQ3}): Can our proposed \emph{PS-Transformer} more accurately select the personality used to generate the response, compared to other baselines? (See Section~\ref{sec:rq3})
    \item (\textbf{RQ4}): What is the impact of the key components in the \emph{PS-Transformer} on performance? (See Section~\ref{sec:rq4})
    \item (\textbf{RQ5}): Can \emph{PS-Transformer}'s performance on IT-ConvAI2 be generalized to the original ConvAI2? (See Section~\ref{sec:rq5})
    \item (\textbf{RQ6}): How does our response method differ from baselines? (See Section~\ref{sec:rq6})
\end{itemize}

\subsection{Datasets}\label{sec:dataset}

\textbf{ConvAI2}\footnote{ConvAI2 is available at \url{https://github.com/facebookresearch/ParlAI/tree/master/parlai/tasks/convai2}.} It is published for the second Conversational Intelligence Challenge~\cite{dinan2020second}, and both speakers of each conversation consist of at least five persona descriptions. The dataset contains 17,878/1,000 multi-turn dialogues conditioned on 1,155/100 personas for train/test. 

\textbf{Inadequate-Tiny-ConvAI2 (IT-ConvAI2)} 
Since ConvAI2 encourages conversation participants to exchange their persona information, speakers tend to express their personas actively without being asked, resulting in fewer OOP queries than in actual practice. To obtain a realistic evaluation of persona-missing conversation, we build IT-ConvAI2 in two steps: 
(1) We first extract queries asking for persona and responses related to personas, respectively. If the extracted query and response are present in a conversation triad $(query,response,personas)$, we will collect them into Tiny-ConvAI2.
(2) To build IT-ConvAI2, for each conversation in Tiny-ConvAI2, those personas involved in response will be removed.
As a result, we manually collect 1,595 conversations as IT-ConvAI2.

\subsection{Baseline Methods}

We compared our proposed approach with the following strong models:

\begin{itemize}
  \item  Generative Based: 
  \textbf{Seq2Seq}~\cite{zhang2018personalizing} is a traditional LSTM-based encoder-decoder model prepending all personas to the query.  
  \textbf{PerCVAE}~\cite{song2019exploiting} further incorporates personas with contexts by a memory network.
  \textbf{DialogWAE}~\cite{gu2019dialogwae} contains a conditional Wasserstein Auto-Encoder, and we adapt it to personalized dialogue generation by concatenating personas with the query directly.
  \item Pre-training \& Fine-tuning Based: 
  \textbf{Transformer}~\cite{dinan2020second} achieves state-of-the-art performance in the manual metrics of the ConvAI2 competition while \textbf{TransferTransfo}~\cite{wolf2019transfertransfo} tops automatic evaluations. 
  \textbf{BERT-Over-BERT (BoB)}~\cite{song2021bob} contains two decoders pretrained on NLI task. It is good at generating responses entailed with personas.
\end{itemize}

\subsection{Evaluation Metrics} \label{sec:metrics}

\subsubsection{Automatic Evaluation.} 
To highlight the quality of generation on both personality and contextual aspects, 
we evaluate each response with two aspects:
\begin{itemize}
\item \textbf{Consistency}: 
Following \citet{dziri2019evaluating}, we employ ESIM\footnote{We use an NLI model different from the one in \emph{PRM} for a fair evaluation.}~\cite{chen2017enhanced} to automatically evaluate the \textbf{entailment score} between the generated response $\mathcal{R}$ and the agent's personas $\mathcal{P} = \{p_1,p_2,...,p_n\}$:
\begin{equation}
\begin{aligned}
e' & = \mathrm{Entail}'(\mathcal{P}, \mathcal{R}) \\
& = \max\limits_{p_i \in \mathcal{P}}\{\mathrm{Entail}'(p_i, \mathcal{R})\}
\end{aligned}
\end{equation}
\item \textbf{Quality}: 
\textbf{BLEU}~\cite{papineni2002bleu} and \textbf{ROUGE}~\cite{lin2004rouge} are used to measure the relevance between the ground truth and generated response. We also employ \textbf{CIDEr}~\cite{vedantam2015cider} to capture the overlap of persona information between the machine response and human reference.
\end{itemize}

\subsubsection{Human Evaluation for PRM} 
Three masters students in the field of dialogue were asked to evaluate per \emph{PRM} according to three metrics:
\begin{itemize}
\item \textbf{Query-relevance ${S_{q}^p}$ (0-1)}: To indicate if the retrieved persona is related to the query based on 1/0 scoring schema.
\item \textbf{Persona-entailment ${S_{p}^p}$ (0-2)}: Scoring how the retrieved persona entails with the query. 0 means conflict, 1 means neutral and 2 means entailment.
\item \textbf{DCG@3}: We collect the top three retrieved results for each method and calculate the DCG@3 in Equation~\ref{eq:dcg}.
\begin{equation}\label{eq:dcg}
\mathrm{DCG}_{3}=\sum_{i=1}^{3} \frac{2^{rel_{i}}-1}{\log _{2}(i+1)}
\end{equation}
where $rel_{i} = \bm{S_{p}^p}$ if the retrieved persona is related to the query, otherwise $rel_{i} = 0$.
\end{itemize}

\subsubsection{Human Evaluation for PS-Transformer.} 
Three judges are asked to evaluate Query-relevance $S_{q}$ (1-3), Persona-entailment $S_{p}$ (1-3) and Response-fluency $S_{r}$ (1-3) of generated responses:
\begin{itemize} 
  \item For \textbf{Query-relevance $S_{q}$}, 1 point means that the response is irrelevant with the query. 2 point means that the response is relevant with query, but is the general response. 3 means that the response perfectly answers the query. 
  \item \textbf{Persona-entailment $S_{p}$} measures whether the response is entailed with predefined personas. 1 means the response doesn't contain any persona. 2 means the response contains persona but not in predefined persona set. 3 means the response contains predefined persona. 
  \item \textbf{Response-fluency $S_{r}$} is used to evaluate the syntactic and logical fluency of the response. The higher the score, the better the performance. 3 point means that the response is both grammatically and logically correct.
\end{itemize}

\subsection{Implementation Details}\label{sec:training}
\begin{itemize}
\item The sentence-pair classifier and the NLI scorer of \emph{PRM} are both BERT-based models. We manually annotate one thousand related $(query, persona)$ pairs for training the sentence pair classifier. NLI scorer is pretrained on both SNLI\footnote{The SNLI is available at \url{https://nlp.stanford.edu/projects/snli}.} and MultiNLI\footnote{The MultiNLI is available at \url{https://cims.nyu.edu/~sbowman/multinli}.}\cite{williams2017broad}, then is finetuned on DNLI\footnote{The DNLI is available at \url{https://wellecks.github.io/dialogue_nli}.}.

\item To train the \emph{Target-Guided Persona Scorer}, we follow \citet{song2019exploiting} labelling each response with its corresponding persona by inverse document frequency. The response has a tf-idf similarity with each persona, and we label each $(response, persona)$ pair with 1/0 according to whether the similarity is higher than a threshold.

\item We employ OpenAI's GPT~\cite{radford2018improving} to initialize \emph{Transformer}, \emph{TransferTransfo} and our \emph{PS-Transformer}. 
\end{itemize}

\section{Results and Analysis}\label{sec:result}

\subsection{Automatic Evaluations (RQ1)}\label{sec:rq1}

As shown in \tabref{tab:result_inde}, we evaluate all the methods on \emph{IT-ConvAI2}, and \emph{IT-ConvAI2 with PRM}, respectively, and we have the following observations:

\textbf{Our pipeline method has the best overall performance in response to OOP queries.} \emph{PS-Transformer} outperforms all baselines regardless of whether our proposed \emph{PRM} is applied to baselines or not. In particular, for \emph{PS-Transformer}, the personality of \emph{PRM} retrieval brings a significant improvement to the entailment of the response. This result shows that in the view of the generative model, the \emph{PRM} retrieval results are very suitable for responding to OOP queries.

\textbf{The \emph{Persona Retrieval Model (PRM)} helps almost all methods generate personality-consistent responses.} 
All methods except \emph{Transformer} reach a higher entailment score after being given a new persona by \emph{PRM}.
Thus, we can generalize a general conclusion that retrieving a suitable persona using our proposed \emph{PRM} in response to an OOP query can help the vast majority of generators to produce a personality-consistent response. It also helps to reduce the risk of fabricating a random persona and generating personality-conflicting responses. In addition, only \emph{PS-Transformer}, when combined with \emph{PRM}, shows a significant improvement not only in entailment but also in response generation quality, which implies that \emph{PS-Transformer} is better than baselines for selection and utilization of personality information.

\subsection{Human Evaluations for \emph{PRM}s (RQ2)}\label{sec:rq2}

\begin{table}[!t]
  \centering
\captionsetup{justification=justified}
\caption{The left shows human evaluation of \emph{PRM}s. The Fleiss´kappa values of $S_q^p, S_p^p$, DCG@3 are 0.62, 0.49, and 0.57, respectively, indicating \emph{Substantial}, \emph{Moderate}, and \emph{Moderate agreement}. The maximum value of the $S_q^p, S_p^p$ are 1, 2, respectively. The right shows automatic evaluations for generated responses based on different \emph{PRM}s.}
  \setlength{\tabcolsep}{4pt}{
    \begin{tabular}{cccccccc}
    \toprule
    \multirow{2}[4]{*}{\textbf{Model}} & \multicolumn{3}{c}{\textbf{Retrieval Quality}} &       & \multicolumn{3}{c}{\textbf{Response Quality}} \\
\cmidrule{2-4}\cmidrule{6-8}          & $S_q^p$  & $S_p^p$  & DCG@3 &       & Entail & Conflict & BLEU \\
    \midrule
    BM25  & 0.35  & 0.86  & 0.77  &       & 0.592 & 0.237 & 5.65 \\
    \midrule
    Classify\textsubscript{SP} & 0.56  & 0.87  & 1.01  &       & 0.650 & 0.304 & 7.16 \\
    \midrule
    NLI\textsubscript{HR} & 0.55  & 0.90  & 1.15  &       & 0.643 & 0.241 & 7.35 \\
    \textbf{NLI\textsubscript{WC}} & \textbf{0.59} & \textbf{0.97} & \textbf{1.33} &       & \textbf{0.670} & \textbf{0.214} & \textbf{7.35} \\
    \bottomrule
    \end{tabular}%
    }
  \label{tab:retriever}%
\end{table}%

To demonstrate the significance of our retrieval methods in Section~\ref{sec:retriever}, we prepare some \emph{PRM}s based on other retrieval methods for an ablation-like experiment:
(1) \textbf{BM25} algorithm is used to retrieve the persona most similar to the query in lexical.
(2) \textbf{Sentence-pair Classifier (Classify\textsubscript{SP})} only adopts the $r$ score in Equation~\ref{eq:relevance} to rank persona candidates.
We employ judges to evaluate a random sample of 150 items per \emph{PRM} according to metrics mentioned in Section~\ref{sec:metrics}.
We further adopt \emph{PS-Transformer} to generate responses based on extended personas from different \emph{Persona Retrieval Model}s, and we apply automatic evaluations on those responses.

\newcommand{\myprm}{\emph{NLI\textsubscript{WC}}}

\textbf{Our proposed \emph{PRM} can effectively solve the OOP problem, and the NLI contributes to improving retrieval performance.}
As retrieval quality is shown in \tabref{tab:retriever}, 59\% of the personas retrieved by our proposed {\myprm} correspond to OOP queries, and the vast majority of these personalities are non-conflicting with predefined personas. Also, {\myprm} outperforms other \emph{PRM}s in terms of the Query-relevance $S^p_q$, the Persona-entailment $S^p_p$, and the overall ranking performance DCG@3.
Specifically, {\myprm} significantly outperforms \emph{BM25} in terms of $S^p_q$ score, indicating that it is effective to consider the semantic relevance between query and retrieved persona. In addition, {\myprm} significantly outperforms \emph{Classify\textsubscript{SP}} in terms of $S^p_p$ score and DCG@3 score, which indicates that natural language inference can effectively reduce the possibility of conflict between retrieved persona and predefined personas, thus improving the final persona retrieval performance.

\textbf{The better the OOP Problem is solved, the higher the response quality of our proposed pipeline method.}
As response quality is shown in \tabref{tab:retriever},
\emph{BM25} retrieves persona considering text similarity only, which makes the retrieved persona weakly correlated with query and even becomes noise during generation. Therefore, \emph{BM25} produces less improvement in Entail score than other \emph{PRM}s and reduces BLEU score that reflects generative performance.
Since \emph{Classify\textsubscript{SP}} ignores the relevance between retrieved persona and predefined personas, the retrieved persona may conflict with predefined personas. In such a case, no matter which persona the generative model selects, the response will conflict with the existing persona set, resulting in a higher Conflict score. 
Compared to \emph{Classify\textsubscript{SP}}, NLI-based \emph{PRM}s (\emph{NLI\textsubscript{HR}} and \emph{NLI\textsubscript{WC}}) reduce the risk of personality conflicts by considering the NLI relevance from retrieval candidates to predefined personas, responses generated based on such \emph{PRM}s also perform well with higher BLEU scores than other \emph{PRM}s.
The results show that \emph{NLI\textsubscript{WC}} outperforms all other \emph{PRM}s in persona retrieval and is most helpful in improving the Entail score and reducing the Conflict score of generated responses.

\subsection{Human Evaluations for Generative Models (RQ3)}\label{sec:humaneval}\label{sec:rq3}

\begin{table}[t]
  \centering
  \captionsetup{justification=justified}
  \caption{Human evaluation of all the generative methods. The Fleiss´kappa values of $S_{q}, S_{p}, S_{r}$ are 0.56, 0.70, and 0.42, respectively, indicating \emph{Moderate}, \emph{Substantial} and \emph{Moderate agreement}. }
    \begin{tabular}{cccc}
    \toprule
    \textbf{Model} & $S_q$    & $S_p$    & $S_r$ \\
    \midrule
    Seq2Seq & 1.40  & 1.75  & 2.59 \\
    PerCVAE & 1.37  & 1.70  & 2.54 \\
    DialogWAE & 1.25  & 1.24  & 2.32 \\
    \midrule
    Transformer & 1.96  & 1.93  & 2.93 \\
    TransferTransfo & 2.12  & 2.40  & 2.90 \\
    BoB   & 2.16  & 2.39  & 2.89 \\
    \midrule
    \textbf{PS-Transformer} & \textbf{2.24} & \textbf{2.44} & \textbf{2.95} \\
    \bottomrule
    \end{tabular}%
  \label{tab:overallscore}%
\end{table}%

\begin{figure*}[!t]
  \centering
  \begin{minipage}[t]{0.49\textwidth}
  \centering
  \includegraphics[width=0.98\textwidth]{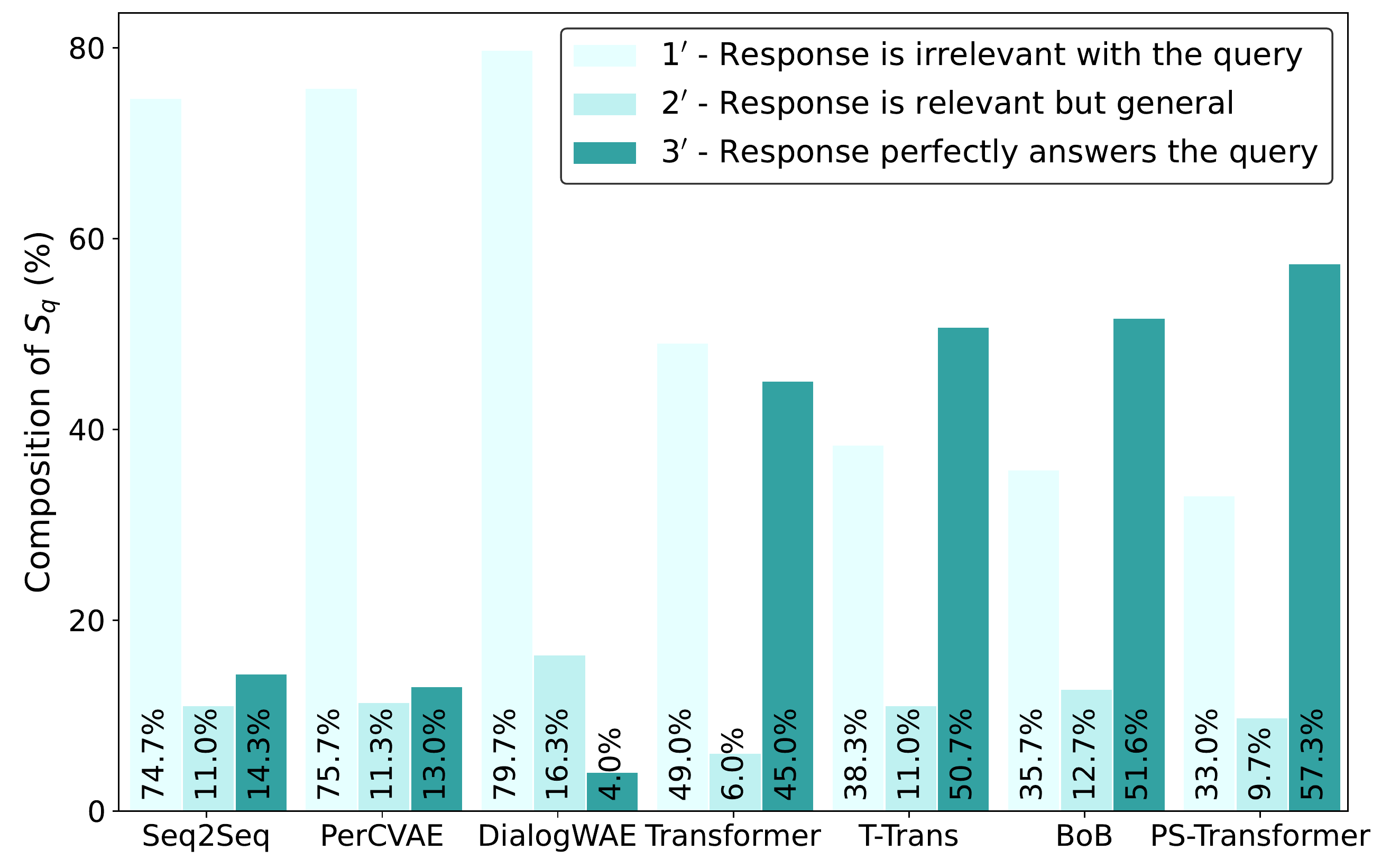}
  \end{minipage}
  \begin{minipage}[t]{0.49\textwidth}
  \centering
  \includegraphics[width=0.98\textwidth]{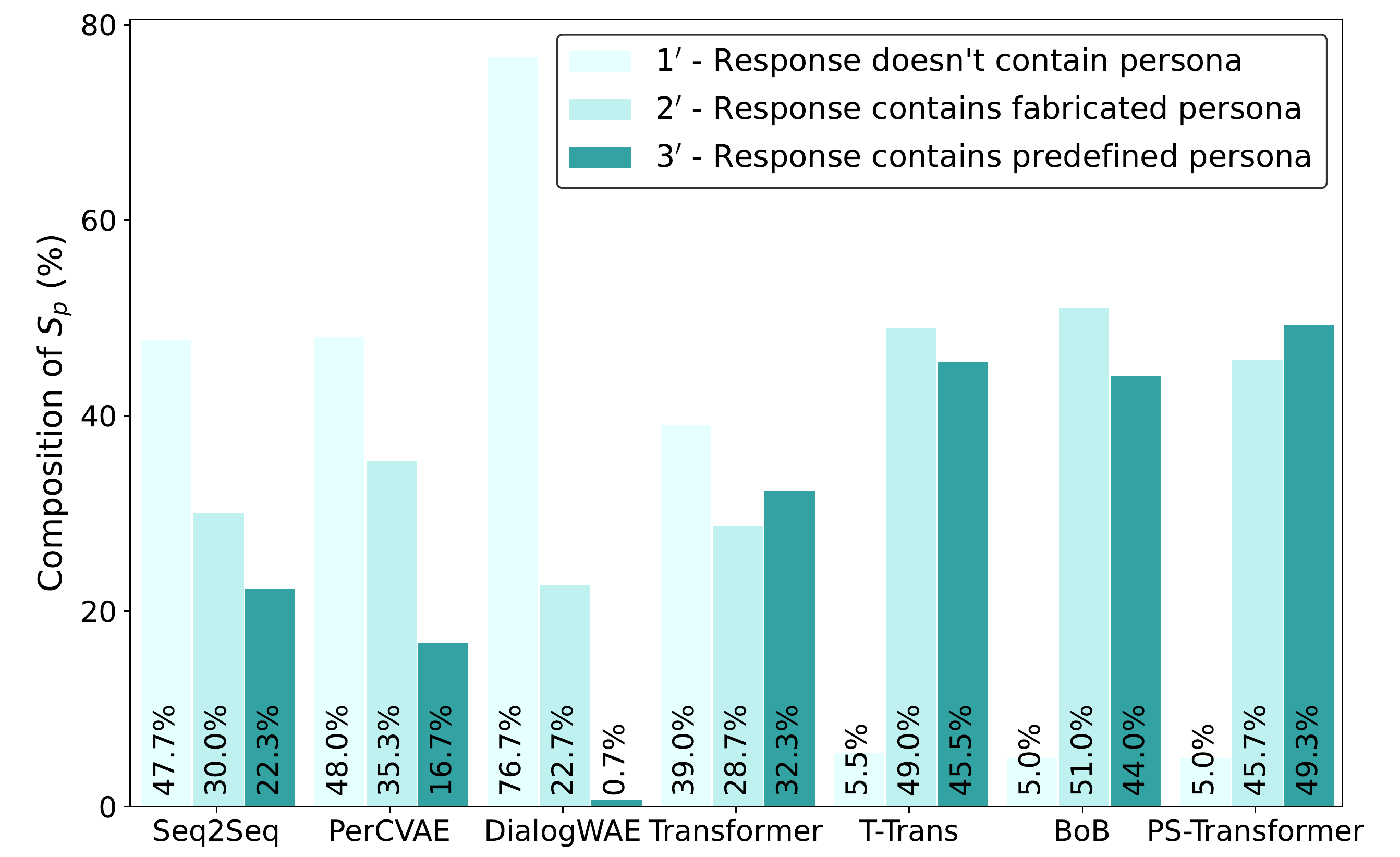}
  \end{minipage}
  \captionsetup{justification=justified}
  \caption{The composition of human evaluation for generated responses (Query-relevance $S_{q}$ and Persona-entailment $S_{p}$). The scoring criteria are shown in the legend. T-Trans is shorthand for TransferTransfo.}
  \label{fig:spe}
\end{figure*}

We randomly sample 150 predictions from column \emph{IT-ConvAI2 with PRM} in \tabref{tab:result_inde} and invite three graduate students for evaluation. The human evaluation results are shown in \tabref{tab:overallscore}, and \figref{fig:spe} shows the detailed compositions of $S_{q}$ and $S_{p}$ scores.

\newcommand{\mygen}{\emph{PS-Transformer}}

\textbf{The \emph{PS-Transformer} outperforms baselines by generating persona-targeted responses.}
As the detailed composition of $S_{p}$ is shown in \figref{fig:spe}, the generated results of {\mygen} have a higher probability of containing predefined personas than all baselines. Compared to \emph{TransferTransfo} and \emph{BoB}, {\mygen}  has a lower probability of fabricating persona. This is because {\mygen} determines which personas should be used before generation, so those selected personas are more likely to be reflected in the response. 
In addition, as the detailed composition of $S_{q}$ is shown in \figref{fig:spe}, responses generated by {\mygen} are the most consistent with queries because the personas selected by {\mygen} in advance are strongly correlated with queries. Thus the responses generated based on the selected personas strongly correlate with the context.
Not only do the results demonstrate that the \emph{Target-Guided Persona Scorer} plays a vital role in accurately selecting persona to generate context-coherence responses,
but they are also consistent with the automatic evaluation result that \emph{PS-Transformer} significantly outperforms other methods in both personality coherence and generating quality.

\subsection{Ablation Study (RQ4)}\label{sec:rq4}

\begin{table}[t]
  \centering
  \caption{Ablation study on \emph{IT-ConvAI2 with PRM}.}
  \setlength{\tabcolsep}{3pt}{
    \begin{tabular}{lccccc}
    \toprule
    \multicolumn{1}{c}{\multirow{2}[4]{*}{\textbf{Settings}}} & \textbf{Consist} &       & \multicolumn{3}{c}{\textbf{Quality}} \\
\cmidrule{2-2}\cmidrule{4-6}          & Entail &       & BLEU  & ROUGE & CIDEr \\
    \midrule
    \textbf{PS-Transformer} & \textbf{0.670} &       & \textbf{7.35} & \textbf{1.73} & \textbf{15.88} \\
    \midrule
    - w/o Posterior Network  & 0.660 &       & 6.71  & 1.64  & 14.67 \\
    - w/o Scorer  & 0.356 &       & 3.54  & 1.02  & 10.05 \\
    \bottomrule
    \end{tabular}%
  }
  \label{tab:ablation}%
\end{table}%

As reported in \tabref{tab:ablation}, we designed and evaluated two variants of \emph{PS-Transformer}: 
(1) We first remove the posterior network (Eq.~\ref{loss:cos}) by directly training the model with prior attention $\mathbf{A}_i^{pri}$. It means we drop the actual personas used in real responses modeled by posterior distribution. It results in deteriorated performance, indicating the importance of the guidance from posterior information.
(2) We remove the entire scoring mechanism (Eq.~\ref{loss:bce}) by treating all personas equally while generating. The significant decrements of all metrics indicate that considering the relevance of personas to query and accurately selecting personas plays an important role in generating personality-consistent and high-quality responses.

\subsection{Effectiveness of \emph{PS-Transformer} on ConvAI2 (RQ5)}\label{sec:rq5}

\begin{table}[!t]
  \centering
  \caption{Automatic evaluation on original ConvAI2.}
    \begin{tabular}{cccccc}
    \toprule
    \multirow{2}[4]{*}{\textbf{Model}} & \textbf{Consist} &       & \multicolumn{3}{c}{\textbf{Quality}} \\
\cmidrule{2-2}\cmidrule{4-6}          & Entail &       & BLEU  & ROUGE & CIDEr \\
    \midrule
    Seq2Seq & 0.092 &       & 5.12  & 1.43  & 9.41 \\
    PerCVAE & 0.287 &       & 2.44  & 0.91  & 5.49 \\
    DialogWAE & 0.047 &       & 3.71  & 1.07  & 5.68 \\
    \midrule
    Transformer & 0.544 &       & 5.47  & 1.37  & 10.87 \\
    TransferTransfo & 0.508 &       & 4.70  & 1.13  & 13.74 \\
    BoB   & 0.499 &       & 4.33  & 1.10  & 9.81 \\
    \midrule
    \textbf{PS-Transformer } & \textbf{0.546} &       & \textbf{6.58} & \textbf{1.49} & \textbf{14.57} \\
    \bottomrule
    \end{tabular}%
  \label{tab:result_ori}%
\end{table}%

\begin{table*}[!htbp]
  \centering
  \caption{Case Study. Responses in red are consistent with agent's personas, and in blue are in conflict with them.}
    \begin{tabular}{cp{4.5cm}p{4.5cm}p{4.5cm}}
    \toprule
    \textbf{Query} & Hello, I'm a art major in college, what do you do? & I am getting in spring, you married? & What grade are you in? \\
    \midrule
    \multirow{3}[2]{*}{\textbf{Predefined Personas}} & I enjoy poetry. & I was poor growing up. & I play with my friends on the playground. \\
    \multicolumn{1}{c}{} & I am a huge star wars fan. & Cheeseburgers are my favorite food. & Mickey mouse is my favorite character. \\
    \multicolumn{1}{c}{} & I played football for a division of a college. & I like watching war documentaries. & I love to go to disney world every year. \\
    \midrule
    \textbf{Extended Persona} & I am studying to be a computer programmer. & I'm married. & I am in the third grade. \\
    \midrule
    Seq2Seq & I am a scientist. I love to write poetry. & No, but i do like watching the flintstones. you? & I am in \textcolor[rgb]{0.2, 0.1, 0.8}{the first grade}. how about you ? \\
    PerCVAE & Who r you on there? & Oh no i live at a beach. & I love to sleep in. \\
    DialogWAE & Do you want to go to the beach? & Wow I love to travel. i love it. & Do have any pets? \\
    Transformer & \textcolor[rgb]{0.2, 0.1, 0.8}{I am a} \textcolor[rgb]{0.9, 0.3, 0.5}{computer science} major at hs. & I am \textcolor[rgb]{0.2, 0.1, 0.8}{getting married in 6 months}. & I am in my \textcolor[rgb]{0.2, 0.1, 0.8}{last year of high school}. \\
    TransferTransfo & \textcolor[rgb]{0.2, 0.1, 0.8}{I am a} \textcolor[rgb]{0.9, 0.3, 0.5}{programmer}. What do you do? & Yes, \textcolor[rgb]{0.9, 0.3, 0.5}{I am married} to my wife. How about you? & My grade is \textcolor[rgb]{0.2, 0.1, 0.8}{5th grade} and you? \\
    BoB   & \textcolor[rgb]{0.2, 0.1, 0.8}{I'm a} \textcolor[rgb]{0.9, 0.3, 0.5}{computer programmer} in college. & No, \textcolor[rgb]{0.2, 0.1, 0.8}{I am not married}. I am a single mom. & I am in \textcolor[rgb]{0.2, 0.1, 0.8}{10th grade}. I love it. \\
    \midrule
    \textbf{PS-Transformer} & I am a student, but I am \textcolor[rgb]{0.9, 0.3, 0.5}{hoping to be a computer science major}. & Yes, \textcolor[rgb]{0.9, 0.3, 0.5}{I'm married} to my wonderful husband. & I am in \textcolor[rgb]{0.9, 0.3, 0.5}{third grade}. \\
    \bottomrule
    \end{tabular}%
  \label{tab:casestudy}%
\end{table*}%

As stated in \tabref{tab:result_ori}, the performance of all the methods on ConvAI2 is consistent with those on IT-ConvAI2 (in \tabref{tab:result_inde}). 
Compared to IT-ConvAI2, a large part of the conversations in ConvAI2 do not even need to be responded to using personas, but our proposed {\mygen} still outperforms all other baselines. The \emph{Target-Guided Persona Scorer} not only selects personas related to the query, but also excludes irrelevant personas as noise, avoiding the deliberate use of personas when generating responses.

\subsection{Case Study (RQ6)}\label{sec:rq6}

In this section, we present an in-depth analysis of response generation of our proposed approach at the level of personality consistency. As shown in \tabref{tab:casestudy}, we prepare three cases generated by different models to explain the superiority of our motivations in personalized dialogue generation.

\textbf{For the first case}: The results suggest that the response generated by our approach is more consistent with personas. 
For instance, the response ``I am a student, but I am hoping to be a computer science major.'' preserves the persona ``to be a programmer''. At the same time, other methods focus on ``programmer'' only.

\textbf{For the second case}: The persona retrieved by \emph{PRM} is proper for the query. The responses generated by \emph{TransferTransfo} and \emph{PS-Transformer} are coherent at both personality and semantic levels when other methods give wrong or irrelevant answers. It should be noted that althought we determine agent's personas as ``married'', it is still possible for agents to fabricate personas about ``gender'', which is a potential problem for further research.

\textbf{For the third case}: The persona retrieved by \emph{PRM} is related to the query and strongly entails all the predefined personas. Though it is hard to exploit persona with numeric information such as ``third grade'' accurately, \emph{PS-Transformer} still generates the response leveraging the proper persona when others give wrong answers.

\subsection{Limitations} \label{sec:limit}

A major limitation of our proposed pipeline is that the global persona set used by \emph{PRM} is constructed in advance, which would make the pipeline still unable to handle OOP queries outside of the entire global persona set. 
A potential solution is introducing a large-scale commonsense knowledge graph (e.g., ConceptNet~\cite{speer2017conceptnet}) to infer new personas, and the utilization of knowledge graphs leaves another research direction.

\section{Conclusion}
In this paper, we propose to tackle the OOP problem in personalized dialogue generation. To tackle the problem above, we formally define the persona extending task and demonstrate that Natural Language Inference can help \emph{PRM} to retrieve a coherent persona for generating response. 
Besides, the \emph{PS-Transformer} introduces a posterior network named \emph{Target-Guided Persona Scorer} that 
help select persona accurately, which help generate personality-consistent responses.
For future work, we will explore how the extended persona affects the next extension to generalize the \emph{retrieval-to-prediction} paradigm over multi-turn conversations.

\section{Acknowledgments}
This work was supported in part by the National Natural Science Foundation of China under Grant No.61602197, Grant No.L1924068, Grant No.61772076, in part by CCF-AFSG Research Fund under Grant No.RF20210005, and in part by the fund of Joint Laboratory of HUST and Pingan Property \& Casualty Research (HPL). 

\bibliographystyle{ACM-Reference-Format}
\balance
\bibliography{cikm22}


\begin{thebibliography}{49}


\ifx \showCODEN    \undefined \def \showCODEN     #1{\unskip}     \fi
\ifx \showDOI      \undefined \def \showDOI       #1{#1}\fi
\ifx \showISBNx    \undefined \def \showISBNx     #1{\unskip}     \fi
\ifx \showISBNxiii \undefined \def \showISBNxiii  #1{\unskip}     \fi
\ifx \showISSN     \undefined \def \showISSN      #1{\unskip}     \fi
\ifx \showLCCN     \undefined \def \showLCCN      #1{\unskip}     \fi
\ifx \shownote     \undefined \def \shownote      #1{#1}          \fi
\ifx \showarticletitle \undefined \def \showarticletitle #1{#1}   \fi
\ifx \showURL      \undefined \def \showURL       {\relax}        \fi
\providecommand\bibfield[2]{#2}
\providecommand\bibinfo[2]{#2}
\providecommand\natexlab[1]{#1}
\providecommand\showeprint[2][]{arXiv:#2}

\bibitem[\protect\citeauthoryear{Bowman, Angeli, Potts, and Manning}{Bowman
  et~al\mbox{.}}{2015}]%
        {bowman2015large}
\bibfield{author}{\bibinfo{person}{Samuel~R. Bowman}, \bibinfo{person}{Gabor
  Angeli}, \bibinfo{person}{Christopher Potts}, {and}
  \bibinfo{person}{Christopher~D. Manning}.} \bibinfo{year}{2015}\natexlab{}.
\newblock \showarticletitle{A large annotated corpus for learning natural
  language inference}. In \bibinfo{booktitle}{\emph{EMNLP}}.
  \bibinfo{publisher}{Association for Computational Linguistics},
  \bibinfo{address}{Lisbon, Portugal}, \bibinfo{pages}{632--642}.
\newblock


\bibitem[\protect\citeauthoryear{Chen, Zhu, Ling, Wei, Jiang, and Inkpen}{Chen
  et~al\mbox{.}}{2017}]%
        {chen2017enhanced}
\bibfield{author}{\bibinfo{person}{Qian Chen}, \bibinfo{person}{Xiaodan Zhu},
  \bibinfo{person}{Zhen-Hua Ling}, \bibinfo{person}{Si Wei},
  \bibinfo{person}{Hui Jiang}, {and} \bibinfo{person}{Diana Inkpen}.}
  \bibinfo{year}{2017}\natexlab{}.
\newblock \showarticletitle{Enhanced {LSTM} for Natural Language Inference}. In
  \bibinfo{booktitle}{\emph{ACL}}. \bibinfo{publisher}{Association for
  Computational Linguistics}, \bibinfo{address}{Vancouver, Canada},
  \bibinfo{pages}{1657--1668}.
\newblock


\bibitem[\protect\citeauthoryear{Dinan, Logacheva, Malykh, Miller, Shuster,
  Urbanek, Kiela, Szlam, Serban, Lowe, Prabhumoye, Black, Rudnicky, Williams,
  Pineau, Burtsev, and Weston}{Dinan et~al\mbox{.}}{2020}]%
        {dinan2020second}
\bibfield{author}{\bibinfo{person}{Emily Dinan}, \bibinfo{person}{Varvara
  Logacheva}, \bibinfo{person}{Valentin Malykh}, \bibinfo{person}{Alexander
  Miller}, \bibinfo{person}{Kurt Shuster}, \bibinfo{person}{Jack Urbanek},
  \bibinfo{person}{Douwe Kiela}, \bibinfo{person}{Arthur Szlam},
  \bibinfo{person}{Iulian Serban}, \bibinfo{person}{Ryan Lowe},
  \bibinfo{person}{Shrimai Prabhumoye}, \bibinfo{person}{Alan~W. Black},
  \bibinfo{person}{Alexander Rudnicky}, \bibinfo{person}{Jason Williams},
  \bibinfo{person}{Joelle Pineau}, \bibinfo{person}{Mikhail Burtsev}, {and}
  \bibinfo{person}{Jason Weston}.} \bibinfo{year}{2020}\natexlab{}.
\newblock \showarticletitle{The Second Conversational Intelligence Challenge
  (ConvAI2)}.
\newblock In \bibinfo{booktitle}{\emph{NeurIPS}},
  \bibfield{editor}{\bibinfo{person}{Sergio Escalera} {and}
  \bibinfo{person}{Ralf Herbrich}} (Eds.). \bibinfo{publisher}{Springer
  International Publishing}, \bibinfo{address}{Cham},
  \bibinfo{pages}{187--208}.
\newblock
\showISBNx{978-3-030-29135-8}


\bibitem[\protect\citeauthoryear{Dziri, Kamalloo, Mathewson, and Zaiane}{Dziri
  et~al\mbox{.}}{2019}]%
        {dziri2019evaluating}
\bibfield{author}{\bibinfo{person}{Nouha Dziri}, \bibinfo{person}{Ehsan
  Kamalloo}, \bibinfo{person}{Kory Mathewson}, {and} \bibinfo{person}{Osmar
  Zaiane}.} \bibinfo{year}{2019}\natexlab{}.
\newblock \showarticletitle{Evaluating Coherence in Dialogue Systems using
  Entailment}. In \bibinfo{booktitle}{\emph{NAACL}}.
  \bibinfo{publisher}{Association for Computational Linguistics},
  \bibinfo{address}{Minneapolis, Minnesota}, \bibinfo{pages}{3806--3812}.
\newblock


\bibitem[\protect\citeauthoryear{Eggins and Slade}{Eggins and Slade}{2005}]%
        {eggins2005analysing}
\bibfield{author}{\bibinfo{person}{Suzanne Eggins} {and} \bibinfo{person}{Diana
  Slade}.} \bibinfo{year}{2005}\natexlab{}.
\newblock \bibinfo{booktitle}{\emph{Analysing casual conversation}}.
\newblock \bibinfo{publisher}{Equinox Publishing}.
\newblock


\bibitem[\protect\citeauthoryear{Fillwock and Traum}{Fillwock and
  Traum}{2018}]%
        {fillwock2018identification}
\bibfield{author}{\bibinfo{person}{Sarah Fillwock} {and} \bibinfo{person}{David
  Traum}.} \bibinfo{year}{2018}\natexlab{}.
\newblock \showarticletitle{Identification of Personal Information Shared in
  Chat-Oriented Dialogue}. In \bibinfo{booktitle}{\emph{LREC}}.
  \bibinfo{publisher}{European Language Resources Association (ELRA)},
  \bibinfo{address}{Miyazaki, Japan}.
\newblock


\bibitem[\protect\citeauthoryear{Gao, Colombo, and Wang}{Gao
  et~al\mbox{.}}{2021}]%
        {gao2021adapting}
\bibfield{author}{\bibinfo{person}{Yang Gao}, \bibinfo{person}{Nicolo Colombo},
  {and} \bibinfo{person}{Wei Wang}.} \bibinfo{year}{2021}\natexlab{}.
\newblock \showarticletitle{Adapting by Pruning: A Case Study on BERT}.
\newblock \bibinfo{journal}{\emph{arXiv:2105.03343}} (\bibinfo{year}{2021}).
\newblock


\bibitem[\protect\citeauthoryear{Gong, Luo, and Zhang}{Gong
  et~al\mbox{.}}{2018}]%
        {gong2018natural}
\bibfield{author}{\bibinfo{person}{Yichen Gong}, \bibinfo{person}{Heng Luo},
  {and} \bibinfo{person}{Jian Zhang}.} \bibinfo{year}{2018}\natexlab{}.
\newblock \showarticletitle{Natural Language Inference over Interaction Space}.
  In \bibinfo{booktitle}{\emph{ICLR}}.
\newblock


\bibitem[\protect\citeauthoryear{Gu, Lu, Li, and Li}{Gu et~al\mbox{.}}{2016}]%
        {gu2016incorporating}
\bibfield{author}{\bibinfo{person}{Jiatao Gu}, \bibinfo{person}{Zhengdong Lu},
  \bibinfo{person}{Hang Li}, {and} \bibinfo{person}{Victor~O.K. Li}.}
  \bibinfo{year}{2016}\natexlab{}.
\newblock \showarticletitle{Incorporating Copying Mechanism in
  Sequence-to-Sequence Learning}. In \bibinfo{booktitle}{\emph{ACL}}.
  \bibinfo{publisher}{Association for Computational Linguistics},
  \bibinfo{address}{Berlin, Germany}, \bibinfo{pages}{1631--1640}.
\newblock


\bibitem[\protect\citeauthoryear{Gu, Ling, Zhu, and Liu}{Gu
  et~al\mbox{.}}{2019b}]%
        {gu2019dually}
\bibfield{author}{\bibinfo{person}{Jia-Chen Gu}, \bibinfo{person}{Zhen-Hua
  Ling}, \bibinfo{person}{Xiaodan Zhu}, {and} \bibinfo{person}{Quan Liu}.}
  \bibinfo{year}{2019}\natexlab{b}.
\newblock \showarticletitle{Dually Interactive Matching Network for
  Personalized Response Selection in Retrieval-Based Chatbots}. In
  \bibinfo{booktitle}{\emph{EMNLP-IJCNLP}}. \bibinfo{publisher}{Association for
  Computational Linguistics}, \bibinfo{address}{Hong Kong, China},
  \bibinfo{pages}{1845--1854}.
\newblock


\bibitem[\protect\citeauthoryear{Gu, Liu, Ling, Liu, Chen, and Zhu}{Gu
  et~al\mbox{.}}{2021}]%
        {gu2021partner}
\bibfield{author}{\bibinfo{person}{Jia-Chen Gu}, \bibinfo{person}{Hui Liu},
  \bibinfo{person}{Zhen-Hua Ling}, \bibinfo{person}{Quan Liu},
  \bibinfo{person}{Zhigang Chen}, {and} \bibinfo{person}{Xiaodan Zhu}.}
  \bibinfo{year}{2021}\natexlab{}.
\newblock \showarticletitle{Partner Matters! An Empirical Study on Fusing
  Personas for Personalized Response Selection in Retrieval-Based Chatbots}. In
  \bibinfo{booktitle}{\emph{SIGIR}} (Virtual Event, Canada)
  \emph{(\bibinfo{series}{SIGIR '21})}. \bibinfo{publisher}{Association for
  Computing Machinery}, \bibinfo{address}{New York, NY, USA},
  \bibinfo{pages}{565–574}.
\newblock
\showISBNx{9781450380379}


\bibitem[\protect\citeauthoryear{Gu, Cho, Ha, and Kim}{Gu
  et~al\mbox{.}}{2019a}]%
        {gu2019dialogwae}
\bibfield{author}{\bibinfo{person}{Xiaodong Gu}, \bibinfo{person}{Kyunghyun
  Cho}, \bibinfo{person}{Jung~Woo Ha}, {and} \bibinfo{person}{Sunghun Kim}.}
  \bibinfo{year}{2019}\natexlab{a}.
\newblock \showarticletitle{Dialogwae: Multimodal response generation with
  conditional Wasserstein auto-encoder}. In \bibinfo{booktitle}{\emph{ICLR}}.
\newblock


\bibitem[\protect\citeauthoryear{Hochreiter and Schmidhuber}{Hochreiter and
  Schmidhuber}{1997}]%
        {hochreiter1997long}
\bibfield{author}{\bibinfo{person}{Sepp Hochreiter} {and}
  \bibinfo{person}{J{\"u}rgen Schmidhuber}.} \bibinfo{year}{1997}\natexlab{}.
\newblock \showarticletitle{Long short-term memory}.
\newblock \bibinfo{journal}{\emph{Neural computation}} \bibinfo{volume}{9},
  \bibinfo{number}{8} (\bibinfo{year}{1997}), \bibinfo{pages}{1735--1780}.
\newblock


\bibitem[\protect\citeauthoryear{Kim, Kang, and Kwak}{Kim
  et~al\mbox{.}}{2019}]%
        {kim2019semantic}
\bibfield{author}{\bibinfo{person}{Seonhoon Kim}, \bibinfo{person}{Inho Kang},
  {and} \bibinfo{person}{Nojun Kwak}.} \bibinfo{year}{2019}\natexlab{}.
\newblock \showarticletitle{Semantic Sentence Matching with Densely-Connected
  Recurrent and Co-Attentive Information}. In \bibinfo{booktitle}{\emph{AAAI}},
  Vol.~\bibinfo{volume}{33}. \bibinfo{pages}{6586--6593}.
\newblock


\bibitem[\protect\citeauthoryear{Li, Galley, Brockett, Spithourakis, Gao, and
  Dolan}{Li et~al\mbox{.}}{2016}]%
        {li2016persona}
\bibfield{author}{\bibinfo{person}{Jiwei Li}, \bibinfo{person}{Michel Galley},
  \bibinfo{person}{Chris Brockett}, \bibinfo{person}{Georgios Spithourakis},
  \bibinfo{person}{Jianfeng Gao}, {and} \bibinfo{person}{Bill Dolan}.}
  \bibinfo{year}{2016}\natexlab{}.
\newblock \showarticletitle{A Persona-Based Neural Conversation Model}. In
  \bibinfo{booktitle}{\emph{ACL}}. \bibinfo{publisher}{Association for
  Computational Linguistics}, \bibinfo{address}{Berlin, Germany},
  \bibinfo{pages}{994--1003}.
\newblock


\bibitem[\protect\citeauthoryear{Lian, Xie, Wang, Peng, and Wu}{Lian
  et~al\mbox{.}}{2019}]%
        {lian2019learning}
\bibfield{author}{\bibinfo{person}{Rongzhong Lian}, \bibinfo{person}{Min Xie},
  \bibinfo{person}{Fan Wang}, \bibinfo{person}{Jinhua Peng}, {and}
  \bibinfo{person}{Hua Wu}.} \bibinfo{year}{2019}\natexlab{}.
\newblock \showarticletitle{Learning to Select Knowledge for Response
  Generation in Dialog Systems}. In \bibinfo{booktitle}{\emph{IJCAI}}.
  \bibinfo{publisher}{International Joint Conferences on Artificial
  Intelligence Organization}, \bibinfo{pages}{5081--5087}.
\newblock


\bibitem[\protect\citeauthoryear{Lin}{Lin}{2004}]%
        {lin2004rouge}
\bibfield{author}{\bibinfo{person}{Chin-Yew Lin}.}
  \bibinfo{year}{2004}\natexlab{}.
\newblock \showarticletitle{{ROUGE}: A Package for Automatic Evaluation of
  Summaries}. In \bibinfo{booktitle}{\emph{Text Summarization Branches Out}}.
  \bibinfo{publisher}{Association for Computational Linguistics},
  \bibinfo{address}{Barcelona, Spain}, \bibinfo{pages}{74--81}.
\newblock


\bibitem[\protect\citeauthoryear{Liu, Mao, Guo, Zhu, Zhou, Hu, and Feng}{Liu
  et~al\mbox{.}}{2021}]%
        {target2021wei}
\bibfield{author}{\bibinfo{person}{Jiayi Liu}, \bibinfo{person}{Xianling Mao},
  \bibinfo{person}{Guibing Guo}, \bibinfo{person}{Feida Zhu},
  \bibinfo{person}{Pan Zhou}, \bibinfo{person}{Yuchong Hu}, {and}
  \bibinfo{person}{Shanshan Feng}.} \bibinfo{year}{2021}\natexlab{}.
\newblock \showarticletitle{Target-Guided Emotion-Aware Chat Machine}.
\newblock \bibinfo{journal}{\emph{ACM Trans. Inf. Syst.}} \bibinfo{volume}{39},
  \bibinfo{number}{4}, Article \bibinfo{articleno}{43} (\bibinfo{date}{aug}
  \bibinfo{year}{2021}), \bibinfo{numpages}{24}~pages.
\newblock
\showISSN{1046-8188}


\bibitem[\protect\citeauthoryear{Liu, Chen, Chen, Lou, Chen, Zhou, and
  Zhang}{Liu et~al\mbox{.}}{2020}]%
        {liu2020you}
\bibfield{author}{\bibinfo{person}{Qian Liu}, \bibinfo{person}{Yihong Chen},
  \bibinfo{person}{Bei Chen}, \bibinfo{person}{Jian-Guang Lou},
  \bibinfo{person}{Zixuan Chen}, \bibinfo{person}{Bin Zhou}, {and}
  \bibinfo{person}{Dongmei Zhang}.} \bibinfo{year}{2020}\natexlab{}.
\newblock \showarticletitle{You Impress Me: Dialogue Generation via Mutual
  Persona Perception}. In \bibinfo{booktitle}{\emph{ACL}}.
  \bibinfo{publisher}{Association for Computational Linguistics},
  \bibinfo{address}{Online}, \bibinfo{pages}{1417--1427}.
\newblock


\bibitem[\protect\citeauthoryear{Majumder, Jhamtani, Berg-Kirkpatrick, and
  McAuley}{Majumder et~al\mbox{.}}{2020}]%
        {majumder2020like}
\bibfield{author}{\bibinfo{person}{Bodhisattwa~Prasad Majumder},
  \bibinfo{person}{Harsh Jhamtani}, \bibinfo{person}{Taylor Berg-Kirkpatrick},
  {and} \bibinfo{person}{Julian McAuley}.} \bibinfo{year}{2020}\natexlab{}.
\newblock \showarticletitle{Like hiking? You probably enjoy nature:
  Persona-grounded Dialog with Commonsense Expansions}. In
  \bibinfo{booktitle}{\emph{EMNLP}}. \bibinfo{publisher}{Association for
  Computational Linguistics}, \bibinfo{address}{Online},
  \bibinfo{pages}{9194--9206}.
\newblock


\bibitem[\protect\citeauthoryear{Mazar{\'e}, Humeau, Raison, and
  Bordes}{Mazar{\'e} et~al\mbox{.}}{2018}]%
        {mazare2018training}
\bibfield{author}{\bibinfo{person}{Pierre-Emmanuel Mazar{\'e}},
  \bibinfo{person}{Samuel Humeau}, \bibinfo{person}{Martin Raison}, {and}
  \bibinfo{person}{Antoine Bordes}.} \bibinfo{year}{2018}\natexlab{}.
\newblock \showarticletitle{Training Millions of Personalized Dialogue Agents}.
  In \bibinfo{booktitle}{\emph{EMNLP}}. \bibinfo{publisher}{Association for
  Computational Linguistics}, \bibinfo{address}{Brussels, Belgium},
  \bibinfo{pages}{2775--2779}.
\newblock


\bibitem[\protect\citeauthoryear{Mitsuda, Higashinaka, and Matsuo}{Mitsuda
  et~al\mbox{.}}{2019}]%
        {mitsuda2019information}
\bibfield{author}{\bibinfo{person}{Koh Mitsuda}, \bibinfo{person}{Ryuichiro
  Higashinaka}, {and} \bibinfo{person}{Yoshihiro Matsuo}.}
  \bibinfo{year}{2019}\natexlab{}.
\newblock \showarticletitle{What Information Should a Dialogue System
  Understand?: Collection and Analysis of Perceived Information in
  Chat-Oriented Dialogue}.
\newblock In \bibinfo{booktitle}{\emph{IWSDS}},
  \bibfield{editor}{\bibinfo{person}{Maxine Eskenazi},
  \bibinfo{person}{Laurence Devillers}, {and} \bibinfo{person}{Joseph Mariani}}
  (Eds.). \bibinfo{publisher}{Springer International Publishing},
  \bibinfo{address}{Cham}, \bibinfo{pages}{27--36}.
\newblock
\showISBNx{978-3-319-92108-2}


\bibitem[\protect\citeauthoryear{Pan, Wei, and Mao}{Pan et~al\mbox{.}}{2021}]%
        {pan2021context}
\bibfield{author}{\bibinfo{person}{Weiran Pan}, \bibinfo{person}{Wei Wei},
  {and} \bibinfo{person}{Xian-Ling Mao}.} \bibinfo{year}{2021}\natexlab{}.
\newblock \showarticletitle{Context-aware Entity Typing in Knowledge Graphs}.
  In \bibinfo{booktitle}{\emph{Findings of EMNLP}}. \bibinfo{pages}{1--8}.
\newblock


\bibitem[\protect\citeauthoryear{Papineni, Roukos, Ward, and Zhu}{Papineni
  et~al\mbox{.}}{2002}]%
        {papineni2002bleu}
\bibfield{author}{\bibinfo{person}{Kishore Papineni}, \bibinfo{person}{Salim
  Roukos}, \bibinfo{person}{Todd Ward}, {and} \bibinfo{person}{Wei-Jing Zhu}.}
  \bibinfo{year}{2002}\natexlab{}.
\newblock \showarticletitle{{B}leu: a Method for Automatic Evaluation of
  Machine Translation}. In \bibinfo{booktitle}{\emph{ACL}}.
  \bibinfo{publisher}{Association for Computational Linguistics},
  \bibinfo{address}{Philadelphia, Pennsylvania, USA},
  \bibinfo{pages}{311--318}.
\newblock


\bibitem[\protect\citeauthoryear{Qian, Huang, Zhao, Xu, and Zhu}{Qian
  et~al\mbox{.}}{2018}]%
        {qian2017assigning}
\bibfield{author}{\bibinfo{person}{Qiao Qian}, \bibinfo{person}{Minlie Huang},
  \bibinfo{person}{Haizhou Zhao}, \bibinfo{person}{Jingfang Xu}, {and}
  \bibinfo{person}{Xiaoyan Zhu}.} \bibinfo{year}{2018}\natexlab{}.
\newblock \showarticletitle{Assigning Personality/Profile to a Chatting Machine
  for Coherent Conversation Generation}. In \bibinfo{booktitle}{\emph{IJCAI}}.
  \bibinfo{publisher}{International Joint Conferences on Artificial
  Intelligence Organization}, \bibinfo{pages}{4279--4285}.
\newblock


\bibitem[\protect\citeauthoryear{Radford, Narasimhan, Salimans, and
  Sutskever}{Radford et~al\mbox{.}}{2018}]%
        {radford2018improving}
\bibfield{author}{\bibinfo{person}{Alec Radford}, \bibinfo{person}{Karthik
  Narasimhan}, \bibinfo{person}{Tim Salimans}, {and} \bibinfo{person}{Ilya
  Sutskever}.} \bibinfo{year}{2018}\natexlab{}.
\newblock \showarticletitle{Improving language understanding by generative
  pre-training}.
\newblock  (\bibinfo{year}{2018}).
\newblock


\bibitem[\protect\citeauthoryear{Shang, Lu, and Li}{Shang
  et~al\mbox{.}}{2015}]%
        {shang2015neural}
\bibfield{author}{\bibinfo{person}{Lifeng Shang}, \bibinfo{person}{Zhengdong
  Lu}, {and} \bibinfo{person}{Hang Li}.} \bibinfo{year}{2015}\natexlab{}.
\newblock \showarticletitle{Neural Responding Machine for Short-Text
  Conversation}. In \bibinfo{booktitle}{\emph{IJCNLP}}.
  \bibinfo{publisher}{Association for Computational Linguistics},
  \bibinfo{address}{Beijing, China}, \bibinfo{pages}{1577--1586}.
\newblock


\bibitem[\protect\citeauthoryear{Song, Wang, Zhang, Zhang, and Liu}{Song
  et~al\mbox{.}}{2021}]%
        {song2021bob}
\bibfield{author}{\bibinfo{person}{Haoyu Song}, \bibinfo{person}{Yan Wang},
  \bibinfo{person}{Kaiyan Zhang}, \bibinfo{person}{Wei-Nan Zhang}, {and}
  \bibinfo{person}{Ting Liu}.} \bibinfo{year}{2021}\natexlab{}.
\newblock \showarticletitle{{B}o{B}: {BERT} Over {BERT} for Training
  Persona-based Dialogue Models from Limited Personalized Data}. In
  \bibinfo{booktitle}{\emph{IJCNLP}}. \bibinfo{publisher}{Association for
  Computational Linguistics}, \bibinfo{address}{Online},
  \bibinfo{pages}{167--177}.
\newblock


\bibitem[\protect\citeauthoryear{Song, Zhang, Cui, Wang, and Liu}{Song
  et~al\mbox{.}}{2019}]%
        {song2019exploiting}
\bibfield{author}{\bibinfo{person}{Haoyu Song}, \bibinfo{person}{Wei-Nan
  Zhang}, \bibinfo{person}{Yiming Cui}, \bibinfo{person}{Dong Wang}, {and}
  \bibinfo{person}{Ting Liu}.} \bibinfo{year}{2019}\natexlab{}.
\newblock \showarticletitle{Exploiting Persona Information for Diverse
  Generation of Conversational Responses}. In
  \bibinfo{booktitle}{\emph{IJCAI}}. \bibinfo{pages}{5190--5196}.
\newblock


\bibitem[\protect\citeauthoryear{Song, Zhang, Hu, and Liu}{Song
  et~al\mbox{.}}{2020}]%
        {song2020generating}
\bibfield{author}{\bibinfo{person}{Haoyu Song}, \bibinfo{person}{Wei-Nan
  Zhang}, \bibinfo{person}{Jingwen Hu}, {and} \bibinfo{person}{Ting Liu}.}
  \bibinfo{year}{2020}\natexlab{}.
\newblock \showarticletitle{Generating persona consistent dialogues by
  exploiting natural language inference}. In \bibinfo{booktitle}{\emph{AAAI}},
  Vol.~\bibinfo{volume}{34}. \bibinfo{pages}{8878--8885}.
\newblock


\bibitem[\protect\citeauthoryear{Speer, Chin, and Havasi}{Speer
  et~al\mbox{.}}{2017}]%
        {speer2017conceptnet}
\bibfield{author}{\bibinfo{person}{Robyn Speer}, \bibinfo{person}{Joshua Chin},
  {and} \bibinfo{person}{Catherine Havasi}.} \bibinfo{year}{2017}\natexlab{}.
\newblock \showarticletitle{ConceptNet 5.5: An Open Multilingual Graph of
  General Knowledge}. In \bibinfo{booktitle}{\emph{AAAI}} (San Francisco,
  California, USA). \bibinfo{pages}{4444–4451}.
\newblock


\bibitem[\protect\citeauthoryear{Vaswani, Shazeer, Parmar, Uszkoreit, Jones,
  Gomez, Kaiser, and Polosukhin}{Vaswani et~al\mbox{.}}{2017}]%
        {vaswani2017attention}
\bibfield{author}{\bibinfo{person}{Ashish Vaswani}, \bibinfo{person}{Noam
  Shazeer}, \bibinfo{person}{Niki Parmar}, \bibinfo{person}{Jakob Uszkoreit},
  \bibinfo{person}{Llion Jones}, \bibinfo{person}{Aidan~N Gomez},
  \bibinfo{person}{\L~ukasz Kaiser}, {and} \bibinfo{person}{Illia Polosukhin}.}
  \bibinfo{year}{2017}\natexlab{}.
\newblock \showarticletitle{Attention is All you Need}. In
  \bibinfo{booktitle}{\emph{NeurIPS}},
  \bibfield{editor}{\bibinfo{person}{I.~Guyon}, \bibinfo{person}{U.~Von
  Luxburg}, \bibinfo{person}{S.~Bengio}, \bibinfo{person}{H.~Wallach},
  \bibinfo{person}{R.~Fergus}, \bibinfo{person}{S.~Vishwanathan}, {and}
  \bibinfo{person}{R.~Garnett}} (Eds.), Vol.~\bibinfo{volume}{30}.
  \bibinfo{publisher}{Curran Associates, Inc.}
\newblock


\bibitem[\protect\citeauthoryear{Vedantam, Zitnick, and Parikh}{Vedantam
  et~al\mbox{.}}{2015}]%
        {vedantam2015cider}
\bibfield{author}{\bibinfo{person}{R. Vedantam}, \bibinfo{person}{C. Zitnick},
  {and} \bibinfo{person}{D. Parikh}.} \bibinfo{year}{2015}\natexlab{}.
\newblock \showarticletitle{CIDEr: Consensus-based image description
  evaluation}. In \bibinfo{booktitle}{\emph{CVPR}}. \bibinfo{publisher}{IEEE
  Computer Society}, \bibinfo{address}{Los Alamitos, CA, USA},
  \bibinfo{pages}{4566--4575}.
\newblock
\showISSN{1063-6919}


\bibitem[\protect\citeauthoryear{Vinyals and Le}{Vinyals and Le}{2015}]%
        {vinyals2015neural}
\bibfield{author}{\bibinfo{person}{Oriol Vinyals} {and} \bibinfo{person}{Quoc
  Le}.} \bibinfo{year}{2015}\natexlab{}.
\newblock \showarticletitle{A Neural Conversational Model}.
\newblock \bibinfo{journal}{\emph{arXiv e-prints}} (\bibinfo{year}{2015}),
  \bibinfo{pages}{arXiv--1506}.
\newblock


\bibitem[\protect\citeauthoryear{Wei, Liu, Mao, Guo, Zhu, Zhou, and Hu}{Wei
  et~al\mbox{.}}{2019}]%
        {emotion2019wei}
\bibfield{author}{\bibinfo{person}{Wei Wei}, \bibinfo{person}{Jiayi Liu},
  \bibinfo{person}{Xianling Mao}, \bibinfo{person}{Guibing Guo},
  \bibinfo{person}{Feida Zhu}, \bibinfo{person}{Pan Zhou}, {and}
  \bibinfo{person}{Yuchong Hu}.} \bibinfo{year}{2019}\natexlab{}.
\newblock \showarticletitle{Emotion-Aware Chat Machine: Automatic Emotional
  Response Generation for Human-like Emotional Interaction}. In
  \bibinfo{booktitle}{\emph{CIKM}}. \bibinfo{publisher}{Association for
  Computing Machinery}, \bibinfo{address}{New York, NY, USA},
  \bibinfo{pages}{1401–1410}.
\newblock
\showISBNx{9781450369763}


\bibitem[\protect\citeauthoryear{Welleck, Weston, Szlam, and Cho}{Welleck
  et~al\mbox{.}}{2019}]%
        {welleck2019dialogue}
\bibfield{author}{\bibinfo{person}{Sean Welleck}, \bibinfo{person}{Jason
  Weston}, \bibinfo{person}{Arthur Szlam}, {and} \bibinfo{person}{Kyunghyun
  Cho}.} \bibinfo{year}{2019}\natexlab{}.
\newblock \showarticletitle{Dialogue Natural Language Inference}. In
  \bibinfo{booktitle}{\emph{ACL}}. \bibinfo{publisher}{Association for
  Computational Linguistics}, \bibinfo{address}{Florence, Italy},
  \bibinfo{pages}{3731--3741}.
\newblock


\bibitem[\protect\citeauthoryear{Williams, Nangia, and Bowman}{Williams
  et~al\mbox{.}}{2018}]%
        {williams2017broad}
\bibfield{author}{\bibinfo{person}{Adina Williams}, \bibinfo{person}{Nikita
  Nangia}, {and} \bibinfo{person}{Samuel Bowman}.}
  \bibinfo{year}{2018}\natexlab{}.
\newblock \showarticletitle{A Broad-Coverage Challenge Corpus for Sentence
  Understanding through Inference}. In \bibinfo{booktitle}{\emph{NAACL}}.
  \bibinfo{publisher}{Association for Computational Linguistics},
  \bibinfo{address}{New Orleans, Louisiana}, \bibinfo{pages}{1112--1122}.
\newblock


\bibitem[\protect\citeauthoryear{Wolf, Sanh, Chaumond, and Delangue}{Wolf
  et~al\mbox{.}}{2019}]%
        {wolf2019transfertransfo}
\bibfield{author}{\bibinfo{person}{Thomas Wolf}, \bibinfo{person}{Victor Sanh},
  \bibinfo{person}{Julien Chaumond}, {and} \bibinfo{person}{Clement Delangue}.}
  \bibinfo{year}{2019}\natexlab{}.
\newblock \showarticletitle{Transfertransfo: A transfer learning approach for
  neural network based conversational agents}.
\newblock \bibinfo{journal}{\emph{arXiv:1901.08149}} (\bibinfo{year}{2019}).
\newblock


\bibitem[\protect\citeauthoryear{Xu, Li, Yang, Ren, Ren, Chen, and Ma}{Xu
  et~al\mbox{.}}{2020}]%
        {xu2020neural}
\bibfield{author}{\bibinfo{person}{Minghong Xu}, \bibinfo{person}{Piji Li},
  \bibinfo{person}{Haoran Yang}, \bibinfo{person}{Pengjie Ren},
  \bibinfo{person}{Zhaochun Ren}, \bibinfo{person}{Zhumin Chen}, {and}
  \bibinfo{person}{Jun Ma}.} \bibinfo{year}{2020}\natexlab{}.
\newblock \showarticletitle{A Neural Topical Expansion Framework for
  Unstructured Persona-Oriented Dialogue Generation}.
\newblock  (\bibinfo{year}{2020}), \bibinfo{pages}{2244--2251}.
\newblock


\bibitem[\protect\citeauthoryear{Xu, Liu, Wang, Sun, and Wang}{Xu
  et~al\mbox{.}}{2017}]%
        {xu2017incorporating}
\bibfield{author}{\bibinfo{person}{Zhen Xu}, \bibinfo{person}{Bingquan Liu},
  \bibinfo{person}{Baoxun Wang}, \bibinfo{person}{Chengjie Sun}, {and}
  \bibinfo{person}{Xiaolong Wang}.} \bibinfo{year}{2017}\natexlab{}.
\newblock \showarticletitle{Incorporating loose-structured knowledge into
  conversation modeling via recall-gate LSTM}. In
  \bibinfo{booktitle}{\emph{IJCNN}}. \bibinfo{pages}{3506--3513}.
\newblock


\bibitem[\protect\citeauthoryear{Yavuz, Rastogi, Chao, and Hakkani-Tur}{Yavuz
  et~al\mbox{.}}{2019}]%
        {yavuz2019deepcopy}
\bibfield{author}{\bibinfo{person}{Semih Yavuz}, \bibinfo{person}{Abhinav
  Rastogi}, \bibinfo{person}{Guan-Lin Chao}, {and} \bibinfo{person}{Dilek
  Hakkani-Tur}.} \bibinfo{year}{2019}\natexlab{}.
\newblock \showarticletitle{{D}eep{C}opy: Grounded Response Generation with
  Hierarchical Pointer Networks}. In \bibinfo{booktitle}{\emph{SIGDIAL}}.
  \bibinfo{publisher}{Association for Computational Linguistics},
  \bibinfo{address}{Stockholm, Sweden}, \bibinfo{pages}{122--132}.
\newblock


\bibitem[\protect\citeauthoryear{Zhang, Dinan, Urbanek, Szlam, Kiela, and
  Weston}{Zhang et~al\mbox{.}}{2018}]%
        {zhang2018personalizing}
\bibfield{author}{\bibinfo{person}{Saizheng Zhang}, \bibinfo{person}{Emily
  Dinan}, \bibinfo{person}{Jack Urbanek}, \bibinfo{person}{Arthur Szlam},
  \bibinfo{person}{Douwe Kiela}, {and} \bibinfo{person}{Jason Weston}.}
  \bibinfo{year}{2018}\natexlab{}.
\newblock \showarticletitle{Personalizing Dialogue Agents: {I} have a dog, do
  you have pets too?}. In \bibinfo{booktitle}{\emph{ACL}}.
  \bibinfo{publisher}{Association for Computational Linguistics},
  \bibinfo{address}{Melbourne, Australia}, \bibinfo{pages}{2204--2213}.
\newblock


\bibitem[\protect\citeauthoryear{Zhang, Sun, Galley, Chen, Brockett, Gao, Gao,
  Liu, and Dolan}{Zhang et~al\mbox{.}}{2020}]%
        {zhang2020dialogpt}
\bibfield{author}{\bibinfo{person}{Yizhe Zhang}, \bibinfo{person}{Siqi Sun},
  \bibinfo{person}{Michel Galley}, \bibinfo{person}{Yen-Chun Chen},
  \bibinfo{person}{Chris Brockett}, \bibinfo{person}{Xiang Gao},
  \bibinfo{person}{Jianfeng Gao}, \bibinfo{person}{Jingjing Liu}, {and}
  \bibinfo{person}{Bill Dolan}.} \bibinfo{year}{2020}\natexlab{}.
\newblock \showarticletitle{{DIALOGPT} : Large-Scale Generative Pre-training
  for Conversational Response Generation}. In \bibinfo{booktitle}{\emph{ACL}}.
  \bibinfo{publisher}{Association for Computational Linguistics},
  \bibinfo{address}{Online}, \bibinfo{pages}{270--278}.
\newblock


\bibitem[\protect\citeauthoryear{Zhao, Wei, Ding, and Mao}{Zhao
  et~al\mbox{.}}{2022}]%
        {zhao22multiview}
\bibfield{author}{\bibinfo{person}{Sen Zhao}, \bibinfo{person}{Wei Wei},
  \bibinfo{person}{Zou Ding}, {and} \bibinfo{person}{Xian-Ling Mao}.}
  \bibinfo{year}{2022}\natexlab{}.
\newblock \showarticletitle{Multi-view Intent Disentangle Graph Networks for
  Bundle Recommendation}. In \bibinfo{booktitle}{\emph{AAAI}}.
  \bibinfo{pages}{1--7}.
\newblock


\bibitem[\protect\citeauthoryear{Zhao, Zhao, and Eskenazi}{Zhao
  et~al\mbox{.}}{2017}]%
        {zhao2017learning}
\bibfield{author}{\bibinfo{person}{Tiancheng Zhao}, \bibinfo{person}{Ran Zhao},
  {and} \bibinfo{person}{Maxine Eskenazi}.} \bibinfo{year}{2017}\natexlab{}.
\newblock \showarticletitle{Learning Discourse-level Diversity for Neural
  Dialog Models using Conditional Variational Autoencoders}. In
  \bibinfo{booktitle}{\emph{ACL}}. \bibinfo{publisher}{Association for
  Computational Linguistics}, \bibinfo{address}{Vancouver, Canada},
  \bibinfo{pages}{654--664}.
\newblock


\bibitem[\protect\citeauthoryear{Zheng, Chen, Huang, Liu, and Zhu}{Zheng
  et~al\mbox{.}}{2019}]%
        {zheng2019personalized}
\bibfield{author}{\bibinfo{person}{Yinhe Zheng}, \bibinfo{person}{Guanyi Chen},
  \bibinfo{person}{Minlie Huang}, \bibinfo{person}{Song Liu}, {and}
  \bibinfo{person}{Xuan Zhu}.} \bibinfo{year}{2019}\natexlab{}.
\newblock \showarticletitle{Personalized dialogue generation with diversified
  traits}.
\newblock \bibinfo{journal}{\emph{arXiv:1901.09672}} (\bibinfo{year}{2019}).
\newblock


\bibitem[\protect\citeauthoryear{Zheng, Zhang, Huang, and Mao}{Zheng
  et~al\mbox{.}}{2020}]%
        {zheng2020pre}
\bibfield{author}{\bibinfo{person}{Yinhe Zheng}, \bibinfo{person}{Rongsheng
  Zhang}, \bibinfo{person}{Minlie Huang}, {and} \bibinfo{person}{Xiaoxi Mao}.}
  \bibinfo{year}{2020}\natexlab{}.
\newblock \showarticletitle{A pre-training based personalized dialogue
  generation model with persona-sparse data}. In
  \bibinfo{booktitle}{\emph{AAAI}}, Vol.~\bibinfo{volume}{34}.
  \bibinfo{pages}{9693--9700}.
\newblock


\bibitem[\protect\citeauthoryear{Zhu, Mo, Zhang, Zhu, Peng, and Yang}{Zhu
  et~al\mbox{.}}{2017}]%
        {zhu2017flexible}
\bibfield{author}{\bibinfo{person}{Wenya Zhu}, \bibinfo{person}{Kaixiang Mo},
  \bibinfo{person}{Yu Zhang}, \bibinfo{person}{Zhangbin Zhu},
  \bibinfo{person}{Xuezheng Peng}, {and} \bibinfo{person}{Qiang Yang}.}
  \bibinfo{year}{2017}\natexlab{}.
\newblock \showarticletitle{Flexible end-to-end dialogue system for knowledge
  grounded conversation}.
\newblock \bibinfo{journal}{\emph{arXiv:1709.04264}} (\bibinfo{year}{2017}).
\newblock


\bibitem[\protect\citeauthoryear{Zou, Wei, Mao, Wang, Qiu, Zhu, and Cao}{Zou
  et~al\mbox{.}}{2022}]%
        {zou2021multi}
\bibfield{author}{\bibinfo{person}{Ding Zou}, \bibinfo{person}{Wei Wei},
  \bibinfo{person}{Xian{-}Ling Mao}, \bibinfo{person}{Ziyang Wang},
  \bibinfo{person}{Minghui Qiu}, \bibinfo{person}{Feida Zhu}, {and}
  \bibinfo{person}{Xin Cao}.} \bibinfo{year}{2022}\natexlab{}.
\newblock \showarticletitle{Multi-level Cross-view Contrastive Learning for
  Knowledge-aware Recommender System}. In \bibinfo{booktitle}{\emph{SIGIR}},
  \bibfield{editor}{\bibinfo{person}{Enrique Amig{\'{o}}},
  \bibinfo{person}{Pablo Castells}, \bibinfo{person}{Julio Gonzalo},
  \bibinfo{person}{Ben Carterette}, \bibinfo{person}{J.~Shane Culpepper}, {and}
  \bibinfo{person}{Gabriella Kazai}} (Eds.). \bibinfo{publisher}{{ACM}},
  \bibinfo{pages}{1358--1368}.
\newblock


\end{thebibliography}










\end{document}